\documentclass[acmtog]{acmart}
\acmSubmissionID{171}

\usepackage{amsmath,amsfonts,bm,multirow}

\definecolor{MyDarkBlue}{rgb}{0,0.08,1}
\definecolor{MyDarkGreen}{rgb}{0.02,0.6,0.02}
\definecolor{MyDarkRed}{rgb}{0.8,0.02,0.02}
\definecolor{MyDarkOrange}{rgb}{0.40,0.2,0.02}
\definecolor{MyPurple}{RGB}{111,0,255}
\definecolor{MyRed}{rgb}{1.0,0.0,0.0}
\definecolor{MyGold}{rgb}{0.75,0.6,0.12}
\definecolor{MyDarkgray}{rgb}{0.66, 0.66, 0.66}
\definecolor{MyDarkCyan}{rgb}{0.05, 0.55, 0.45}
\definecolor{MyBlack}{rgb}{0., 0., 0.}
\definecolor{MyMagenta}{rgb}{1., 0., 1.}
\definecolor{BerkeleyYellow}{RGB}{255,204,41}
\definecolor{BerkeleyLightBlue}{RGB}{94,146,221}
\definecolor{BkDarkBlue}{rgb}{.05,.07,.353}

\definecolor{MyDarkGray2}{rgb}{0.6, 0.6, 0.6}

\newcommand{\camready}[1]{#1}
\newcommand{\suppmat}[1]{\textcolor{MyBlack}{#1}}

\def\rvz{{\mathbf{z}}}

\newcommand{\reffig}[1]{Figure~\ref{fig:#1}}
\newcommand{\refsec}[1]{Section~\ref{sec:#1}}

\newcommand{\reftbl}[1]{Table~\ref{tbl:#1}}

\newcommand{\ignorethis}[1]{}

\newcommand{\myparagraph}[1]{\vspace{1pt} \noindent \textbf{#1} \ }

\def\1{\bm{1}}

\newcommand{\latenti}{\latent_{(i)}}
\newcommand{\editop}{\mathcal{T}_{(i)}}
\newcommand{\og}{G(\latent; \theta)}

\newcommand{\ogsp}{G(\latenti; \theta)}
\newcommand{\mgsp}{G(\latenti; \theta')}
\newcommand{\editsp}{\editop(G(\latenti; \theta))}
\newcommand{\tsample}{\{\latenti, \editsp\}_{i=1}^{N}}

\newcommand{\lossproj}{\mathcal{L}_{\text{LPIPS}}}

\newcommand{\latshapei}{\latenti}
\newcommand{\lattexture}{\latent_{\text{t}}}
\newcommand{\ogspsty}{G(\latshapei, \lattexture; \theta)}
\newcommand{\mgspsty}{G(\latshapei, \lattexture; \theta')}
\newcommand{\editspsty}{\editop(\ogspsty)}
\newcommand{\mgspstylayer}{G(\latshapei, \lattexture; W_j')}
\newcommand{\editspstylayer}{\editop(G(\latshapei, \lattexture; W_j))}

\newcommand{\mcolor}{M_{\text{color}}}
\newcommand{\mfixed}{M_{\text{fixed}}}
\newcommand{\lcolor}{\lambda_{\text{color}}}

\newcommand{\latent}{{\rvz}}

\DeclareMathOperator*{\argmin}{arg\,min}

\newcommand{\ignore}[1]{}

\renewcommand*{\thefootnote}{\arabic{footnote}}

\makeatletter
\DeclareRobustCommand\onedot{\futurelet\@let@token\@onedot}
\def\@onedot{\ifx\@let@token.\else.\null\fi\xspace}

\def\etal{\emph{et al}\onedot}
\makeatother

 \usepackage{booktabs} %

\usepackage[ruled]{algorithm2e} %

\SetAlFnt{\small}
\SetAlCapFnt{\small}
\SetAlCapNameFnt{\small}
\SetAlCapHSkip{0pt}

\acmJournal{TOG}

\setcopyright{rightsretained}
\acmJournal{TOG}
\acmYear{2022}\acmVolume{41}\acmNumber{4}\acmArticle{73}\acmMonth{7} \acmDOI{10.1145/3528223.3530065}

\begin{document}
\title{Rewriting Geometric Rules of a GAN}

\author{Sheng-Yu Wang}
\orcid{0000-0003-4000-2046}
\affiliation{%
 \institution{Carnegie Mellon University}
 \streetaddress{5000 Forbes Ave}
 \city{Pittsburgh}
 \state{PA}
 \postcode{15213}
 \country{USA}
}
\email{shengyu2@andrew.cmu.edu}

\author{David Bau}
\orcid{0000-0003-1744-6765}
\affiliation{%
\institution{Northeastern University}
\streetaddress{440 Huntington Ave}
\city{Boston}
\state{MA}
\postcode{02115}
\country{USA}}
\email{davidbau@northeastern.edu}

\author{Jun-Yan Zhu}
\orcid{0000-0001-8504-3410}
\affiliation{%
 \institution{Carnegie Mellon University}
 \streetaddress{5000 Forbes Ave}
 \city{Pittsburgh}
 \state{PA}
 \postcode{15213}
 \country{USA}
}
\email{junyanz@andrew.cmu.edu}

\begin{abstract}
Deep generative models make visual content creation more accessible to novice users by automating the synthesis of diverse, realistic content based on a collected dataset. However, the current machine learning approaches miss a key element of the creative process -- the ability to synthesize things that go far beyond the data distribution and everyday experience. 
To begin to address this issue, we enable a user to \camready{``warp'' a given model by editing just a handful of original model outputs with desired geometric changes. Our method applies a low-rank update to a single model layer to reconstruct edited examples. Furthermore, to combat overfitting, we propose a latent space augmentation method based on style-mixing.} Our method allows a user to create a model that synthesizes endless objects with defined geometric changes, enabling the creation of a new generative model without the burden of curating a large-scale dataset. We also demonstrate that edited models can be composed to achieve aggregated effects, and we present an interactive interface to enable users to create new models through composition. Empirical measurements on multiple test cases suggest the advantage of our method against recent GAN fine-tuning methods.  Finally, we showcase several applications using the edited models, including latent space interpolation and image editing. %
\end{abstract}



\begin{teaserfigure}
    \centering
    \includegraphics[width=\linewidth]{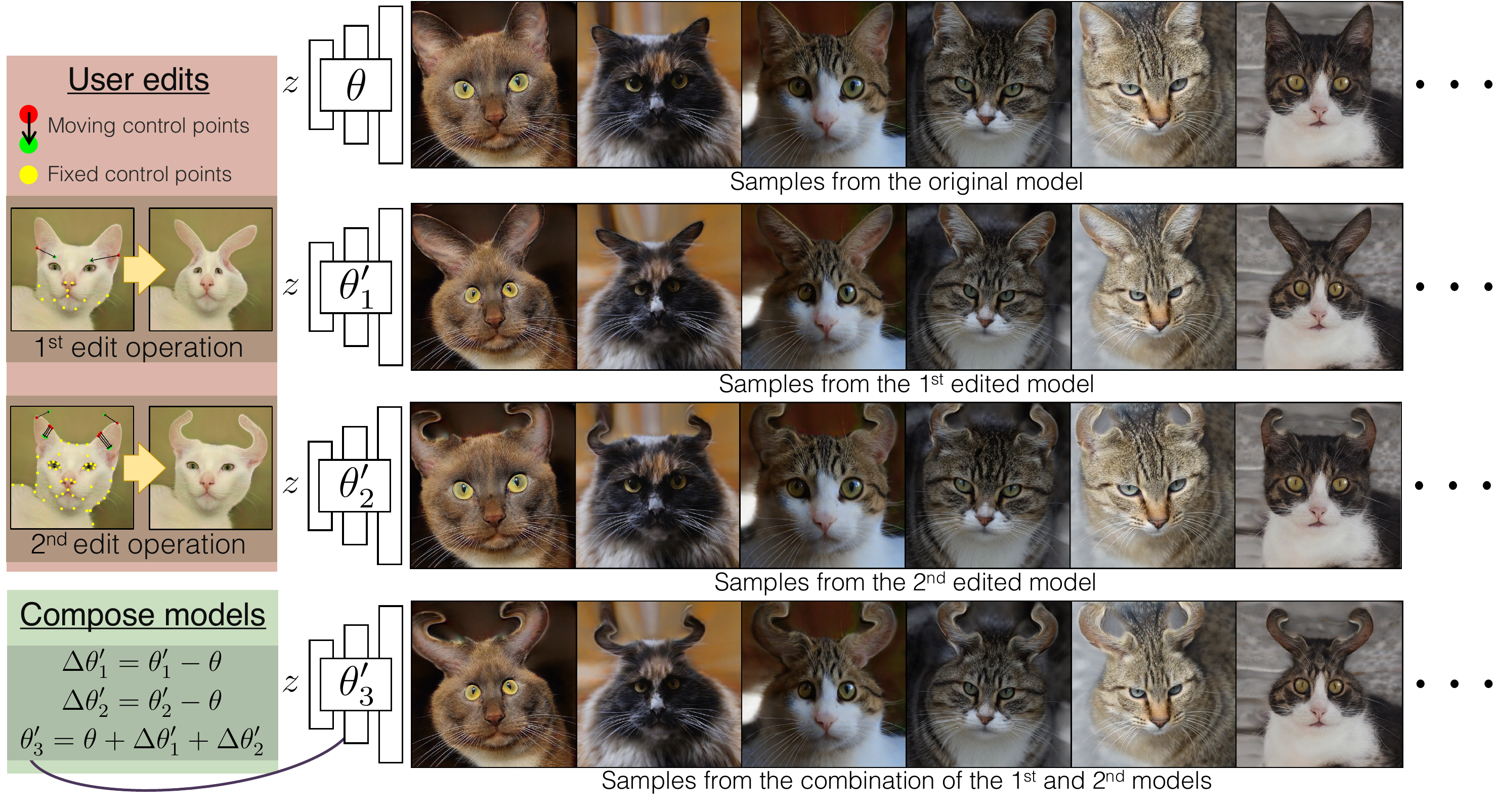}
    \caption{%
    With our method, a user can edit a GAN model to synthesize many unseen objects with the desired shape. The user is asked to warp just a handful of generated images by defining several control points \textbf{(User edits)}, to obtain the customized models \textbf{(1st and 2nd edited models)}. Furthermore, a user can compose the edited models into a new model with aggregated geometric changes \textbf{(last row)}. For each row, the same noise $z$ is used for all models. While the edited models change an object’s shape, other visual cues, such as pose, color, texture, and background, are faithfully preserved after the modification.
    }
    \label{fig:teaser}
\end{teaserfigure}

 \maketitle

\section{Introduction}

Deep generative models have reduced the technical barriers to creating visual content. They can free a person from the need to develop all the skills to create the fine details of a realistic image or an art piece. 
Instead, by taking simple user instructions, generative models apply statistical prowess to synthesize realistic, diverse content and fill in details that imitate observed patterns in our visual world. Using a generative model, a novice user can create a magnificent landscape scene with a few brushstrokes~\cite{park2019SPADE}, synthesize a video of themselves performing expert dance moves~\cite{chan2019dance}, or even generate and manipulate a photo based on a simple text prompt~\cite{ramesh2021zero,patashnik2021styleclip}. 

These applications all use generative models as a data-driven source, so that newly-created content will resemble previous content in the training data distribution. However, we humans do not limit our creativity to imitations of things that already exist. For instance, the creators of movies such as Avatar~\cite{avatar2009} or Star Wars~\cite{starwars1977} imagined fantastical creatures unlike any in the real world, inspiring wonder and excitement about a vivid universe beyond everyday experience. Currently, to create a generative model of such virtual creatures, an artist could fabricate a number of examples from scratch and apply few-shot GAN training methods~\cite{zhao2020diffaugment,li2020fewshot,ojha2021few-shot-gan,karras2020ADA}. However, the process of assembling enough data to apply these methods is still time-consuming, costly, and not an option for everyday users.

This leads to the question: can a novice user apply changes to a generative model to synthesize new content that differs from existing examples? We draw inspiration from the creative process of visual effects artists who imagine new worlds~\cite{failes2016masters}. Designing a new creature can begin by warping an existing species into a completely new shape
to achieve the desired effect. The artist redefines the geometric rules of a creature, for example, re-imagining ordinary shapes of animal eyes into expressive new shapes that portray a particular personality. We ask if such a creative process can be applied to manipulate a generative model, using simple user inputs to change the geometric rules within a learned model.
A manipulated generative model would be able to synthesize a universe of creatures with diversity and variation, all following new geometric rules defined by the artist.

\camready{We present a new problem setting: user-guided modification of the geometric rules encoded in a pre-trained generative model. To edit a model, we ask the users to move a handful of control points as warping examples. Exisiting GAN inversion~\cite{zhu2016generative} and few-shot GAN adaptation~\cite{ojha2021few-shot-gan} methods fail to tackle this new problem: the geometric changes can go beyond the learned distribution, placing the goal out of reach of GAN inversion; and given extremely few user inputs, it is difficult to adapt a GAN to the new domain. To address these issues, we directly optimize the model weights with a reconstruction loss between original samples and their warped versions. We introduce an augmentation scheme based on style-mixing~\cite{karras2019style} to improve our model's generalization to unseen samples, and we show that we can update only a small subset of networks weights to rewrite the geometric rules. Moreover, by a simple linear combination of edited model's weights, we observe that the edited models can be composed to build a model with combined shape-changing effects. Based on this finding, we present an interactive interface that allows users to build new models by composing different models.}

\camready{As shown in \reffig{teaser}, our method enables a user to edit a whole generative model that synthesizes an endless collection of creatures, such as the long-eared rabbit cats, or the devilish cats with curly-shaped ears. We can also combine two warped models to create a new model of rabbit cats with curly ears. }

We use our method to create several edited generative models, and we demonstrate several applications of these models, including smooth transitions between generated images, GAN-based image manipulations, and real photo editing. We also show how to  generalize our formulation to apply coloring changes to a generative model. Furthermore, we evaluate our method on multiple warping operations on different classes to fully characterize its performance. We show that our method outperforms recent few-shot GAN fine-tuning methods. Finally, we include an extensive ablation study regarding several algorithmic choices and hyperparameters.  \camready{
Our code and models are available at our website:  \href{https://peterwang512.github.io/GANWarping}{https://peterwang512.github.io/GANWarping}.}%

\begin{figure*}
    \centering
    \includegraphics[width=0.95\linewidth]{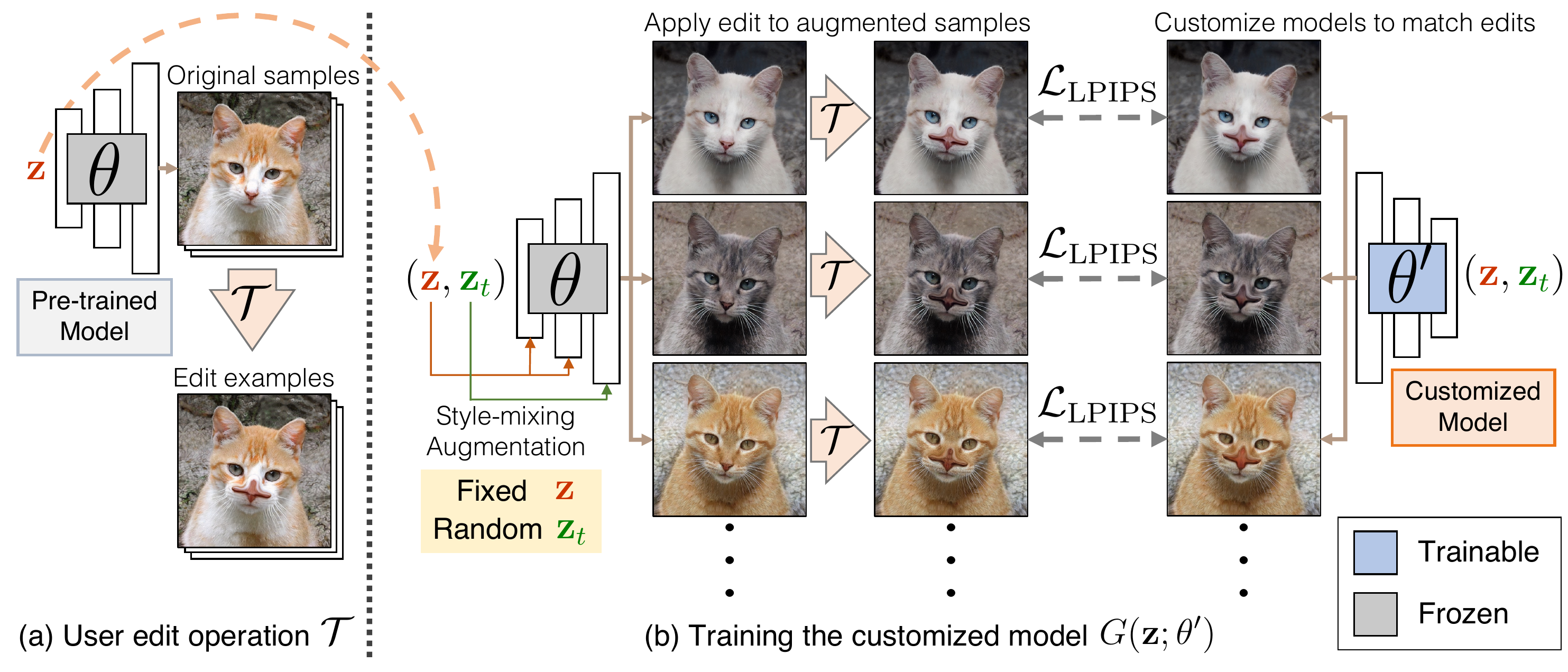}
    \caption{{\bf Overview.} (a) A user first edits a handful of samples from the pre-trained generative model. Each user edit $\editop$ warps the input $\ogsp$ to output $\editsp$.  (b) We train a customized model $\theta'$ so that it can synthesize new samples with a similar visual effect specified by the user edits $\editop$. To prevent overfitting, we apply style-mixing augmentation to the edited samples (\refsec{stylemix}). For each sample, we mix the original latent code $\latenti$ with a new randomly sampled texture latent code $\lattexture$. Since the augmented samples still preserve shapes and poses, we can apply the same user edit $\editop$  to obtain a training set with diverse texture variations. We learn the customized model $\theta'$ on the augmented training set using the LPIPS loss~\cite{zhang2018unreasonable}. In the figure, we denote $\latenti$ by $\mathbf{z}$ and $\editop$ by $\mathcal{T}$ for brevity.}
    \label{fig:method}
\end{figure*}

\section{Related Works}
\label{sec:related}
\paragraph{Generative models for content creation.} Deep generative models~\cite{goodfellow2014generative,dinh2016density,kingma2014auto,oord2016conditional} and its recent advances~\cite{karras2019style,brock2019large,kingma2018glow,Karras2021alias,razavi2019generating,ho2020denoising,gal2021swagan} have enabled a wide range of image and video synthesis applications. Recent applications include text-based image synthesis and editing~\cite{ramesh2021zero,patashnik2021styleclip}, semantic photo synthesis~\cite{park2019SPADE}, motion transfer~\cite{chan2019dance,wang2018vid2vid}, image manipulation~\cite{zhu2016generative,tov2021designing,richardson2021encoding}, virtual Try-on~\cite{lewis2021tryongan,albahar2021pose}, and face editing~\cite{or2020lifespan,portenier2018faceshop,alaluf2021matter,abdal2021styleflow,tan2020michigan}. Generative models excel at  synthesizing realistic samples as they learn the prior of natural image statistics from training data. Unfortunately, they are also confined to imitating the training data, making synthesizing out-of-the-distribution samples incredibly difficult. Different from the above works, our goal is \emph{not} to synthesize or edit an image via a generative model. Instead, we aim to change the geometric rules of the model without curating a large-scale training set. %

\paragraph{Model fine-tuning and rewriting.}
Various methods have been proposed to adapt the pre-trained generative model to a small unseen domain. Fine-tuning from pre-trained weights can improve upon training from scratch~\cite{wang2018transferring} but is also prone to overfitting. To alleviate this issue, several data augmentation methods~\cite{zhao2020image,karras2020ADA,zhao2020diffaugment,tran2020towards} and regularization methods~\cite{li2020fewshot,ojha2021few-shot-gan,lecamgan} are proposed. Some also suggest limiting the changes in model weights~\cite{noguchi2019image,mo2020freeze,zhao2020leveraging,wang_2020_CVPR}, and some use models pre-trained on other vision tasks to make the discriminator more robust~\cite{Sauer2021NEURIPS,sung2018,Gadde2021DetailMM,kumari2021ensembling}. Cross-domain methods fine-tune a generative model to match user-specified sketches~\cite{wang2021sketch} or text prompts~\cite{gal2021stylegan}. Our work differs from these previous approaches since our task does \emph{not} adapt a model to an existing target distribution, and our rewritten rules can apply out-of-distribution shape changes that cannot be achieved by latent manipulations~\cite{zhang2021image,ling2021editgan,Patashnik_2021_ICCV}. %
We enable the user to define a novel target directly through manipulation of the model. Our approach is inspired by
work that rewrites object association rules of a deep network~\cite{bau2020rewriting,santurkar2021editing}, but while those works apply to copy-and-paste changes, our method addresses the new problem of rewriting the geometric rules that define shapes in a generative model.

\paragraph{Image warping.} Image warping is a classic image editing problem in computer graphics and computational photography. There are mainly three types of warping operations. Global warping aims to transform every single pixel via the same transformation matrix and has been widely used in panorama image stitching\cite{brown2007automatic} and  registration~\cite{fischler1981random}. However, global transformation cannot model complex local geometric changes. In contrast, we can use a per-pixel dense flow field to warp an input image into an arbitrary shape, in which each vector in the flow field describes the pixel movement between two images. Thanks to its expressiveness, they have been used in image editing and style transfer~\cite{shih2013data,shih2014style,Liao:2017:VAT:3072959.3073683,barnes2009patchmatch} as well as image alignment~\cite{liu2010sift,lucas1981iterative}. The third approach is to achieve local deformation through sparse correspondence. This approach allows a user to specify a small number of source and target control points or lines, and the algorithm then constructs a dense flow field using various interpolation techniques~\cite{schaefer2006mls,igarashi2005rigid,alexa2000rigid,wood2003thin,kim2019facial}. The reconstructed flow field needs to satisfy the sparse user controls while staying smooth in the spatial domain. In this work, we leverage Moving Least Squares~\cite{schaefer2006mls} to create editing examples. \camready{Recently, conditional generative models such as WarpGAN~\cite{warpgan}, CariGAN~\cite{cao2018cari}, and StyleCariGAN~\cite{Jang2021StyleCari} warp and stylize faces to create caricatures.} However, unlike previous methods, which aim to warp a single image, we aim to change the weights of a neural network so that the resulting model can consistently produce samples that reflect the desired geometric changes without additional user inputs.  %

\section{Methods}

We aim to modify the geometric rules of a generative model according to a few user edits. Given a pre-trained generator $\og$, a user is asked to edit a small number of generated samples. \camready{For generative models that use an intermediate latent space (e.g., StyleGAN3~\cite{Karras2021alias}), we denote the intermediate latent code by $\latent$ to make notations consistent.} $\tsample$ denotes the sampled latent codes and corresponding edits, where $N$ is the number of samples, and $\editop$ is the user-defined local warping function for the $i$-th each generated image. 
We often refer to them as training examples as we train our model with these edits. 
Our goal is to learn a new model $\theta'$, whose samples resemble the visual effect of user edits.

We introduce a simple method to edit the GAN model. In \refsec{loss}, we formulate our training objective and discuss several alternative formulations. 
While directly applying this objective leads to severe overfitting, we drastically improve our model's generalization via a latent space augmentation method based on the disentanglement structure of the generative model (\refsec{stylemix}). Furthermore, we perform model updates on a smaller set of parameters to significantly reduce model storage (\refsec{parameter}). We then demonstrate how to compose multiple geometric changes into a single model and present an interactive interface for users to create their own models (\refsec{compose}). Finally, in \refsec{color}, we extend our method to achieve color changes.

\subsection{Adapting Generator Weights to User Edits}
\label{sec:loss}
\begin{figure}
    \centering
    \includegraphics[width=\linewidth]{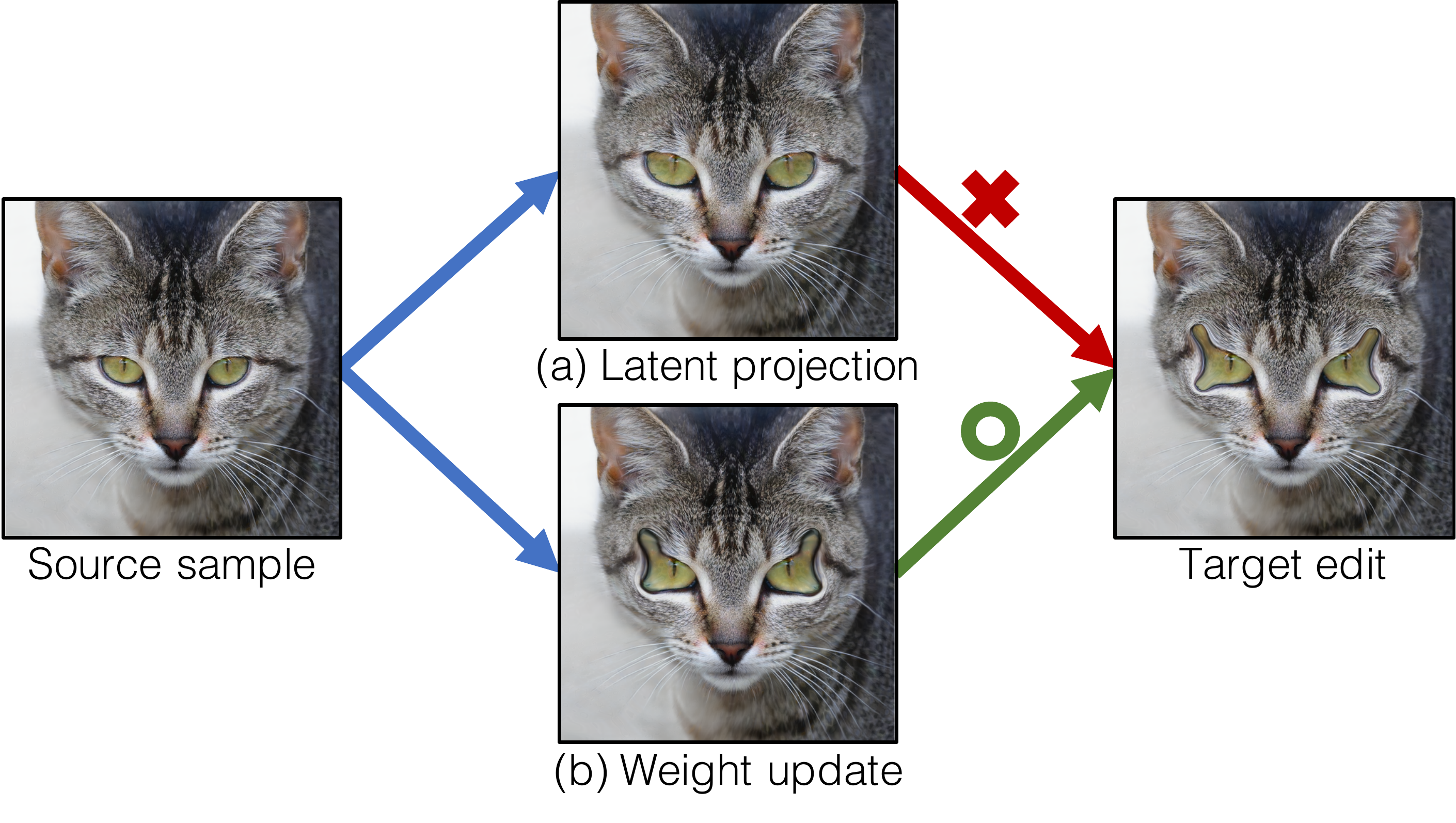}
    \caption{{\bf Reconstructing an image with a drastic warp edit.} We reconstruct a model-generated sample that is warped drastically by (a) projecting the edited image into the extended latent space proposed by Abdal~\etal~\shortcite{abdal2020image2stylegan} or (b) updating the generative model weights directly. The latent projection method can only produce an image with bigger eyes, whereas the weight update method correctly matches the target edits. The latent code from the source sample is used to initialize GAN projection.}
    \label{fig:latent_proj}
\end{figure}

\paragraph{Warping interface for collecting training examples.}
To make model creation accessible to a novice user, we present a warping interface wherein the user can define the warping function $\editop$ intuitively. %
\camready{We create training edit examples using random sampling, and the user can manually reject samples when editing is not feasible (e.g., parts to edit are occluded). For each sample, a user clicks to define a control point p, then drags and releases the click to define the new position q.}
The control points indicate where and how each part of the generated image should move. Given user control points, we compute the image deformation grid and apply inverse warping using Moving Least Squares~\cite{schaefer2006mls} to obtain the edited examples $\editsp$. As shown in \reffig{teaser}, a user can warp a cat face into a rabbit face or deform a cat ear into a curly shape within a few clicks. Thanks to the flexibility of the control points, our users can quickly introduce a wide range of geometric changes. Our method is not tied to a particular local deformation algorithm and can be used with other methods~\cite{wood2003thin,alexa2000rigid,igarashi2005rigid} and software (e.g., Photoshop Liquify). %

\paragraph{Challenges and earlier attempts.} Altering the geometric rules of a model is a challenging task. To create a local geometric change, such as changing the shape of a nose, the new model needs to decide how to move the correct parts to the right place. To achieve this, the model has to detect object parts and understand geometric relations between different parts. However, given a few training examples, it might be difficult to learn these concepts and generalize well to unseen samples. %

In the early stage, we experimented with two problem formulations. Our first attempt was to formulate it as a few-shot GAN training problem~\cite{wang2018transferring,ojha2021few-shot-gan,zhao2020diffaugment,karras2020ADA} and treated the edited images as training data for GANs. Unfortunately, direct GAN training, even with pre-trained generators, regularization, and data augmentation, still fails to work due to the limited number of training samples (i.e., less than 10).  We then cast the problem as an image projection problem~\cite{zhu2016generative,abdal2020image2stylegan,huh2020ganprojection,richardson2020encoding}. Image projection aims to obtain the optimal latent vector such that the generated sample reconstructs the target image. One can further traverse the latent space to manipulate images. However, as demonstrated by \reffig{latent_proj}, it is difficult to even project a single warped output to the latent space.

\paragraph{Our learning objective.} As it is difficult to manipulate shapes by only updating the latent vector,  we propose updating the \textit{network weights} to change the geometric rules inside the models directly. This brings two potential benefits. First, we can reconstruct training samples more easily due to the high-dimensional nature of weight space. Second, we observe that it is easier to introduce out-of-the-distribution geometric changes (e.g., deform a
cat ear into a curly shape) compared to using latent directions~\cite{harkonen2020ganspace}.  

More concretely, we optimize the model weights to reconstruct the edited examples using the following loss:
\begin{equation}
    \argmin_{\theta'} \frac1N \sum_{i=1}^{N} \lossproj(\mgsp, \editsp),
\end{equation}
where $\lossproj$ is the LPIPS loss~\cite{zhang2018unreasonable} that measures the similarity between two images based on deep feature embeddings. The objective above encourages the new model $\theta'$ to mimic the user edits $\editop$ for each generated example $\ogsp$ over N examples. We initialize the weights as $\theta'=\theta$ for faster convergence. Regarding the number of training examples $N$, we have included an ablation study for $N \in \{1, 2, 4, 6, 8, 10\}$. We observe that our method works well for 5 to 10 user edits. %

\subsection{Style-mixing Augmentation}
\label{sec:stylemix}
However, updating weights also introduce a new issue: overfitting. %
While this method accurately reconstructs the training examples, the resulting model weights do not generalize to unseen samples, as shown in \reffig{stylemix_effect}. Collecting more user edits (e.g., hundreds or thousands) can potentially alleviate the overfitting issue. Unfortunately, it requires a significant amount of manual efforts.%

To resolve this, we apply a latent space augmentation scheme that utilizes the disentanglement of the generative model. In state-of-the-art GANs~\cite{brock2019large,Karras2021alias}, the layer weights are conditioned on the latent vectors, and StyleGAN~\cite{karras2019style} observes that the latent codes in the earlier layers control the shape of objects, while those in the later layers control color and texture. Furthermore, one can perform ``style-mixing,'' where two different latent codes are applied to the earlier and later layers respectively. The two different latent codes control different aspects of the sampling process. %

Building on this concept, we can obtain infinitely many new data points from each warping edit by randomly sampling the latent codes of the color and texture. Since the original latent code controlling the shape does not change, the same warping function can be applied to the new samples and still achieve the desired edits. Mathematically speaking, the objective becomes:
\begin{equation}
     \argmin_{\theta'} \frac1N \sum_{i=1}^{N} \mathbb{E}_{\lattexture} \left[ \lossproj(\mgspsty, \editspsty) \right]. 
\end{equation}

As shown in \reffig{method} (b), $\latshapei$ is the original latent code, which we used to create the input example. During training time, we fix $\latshapei$ to preserve the shape of the input image. We then mix it with a randomly sampled latent code $\lattexture$ to introduce new texture variations and create \emph{free} training examples. \reffig{stylemix_effect} shows the comparison between models trained with and without the style-mixing augmentation, and it is evident that the models trained with our augmentation generalize much better. %

\subsection{Updating Models with Fewer Parameters}
\label{sec:parameter}
Inspired by the recent works on fine-tuning generative models~\cite{bau2020rewriting,pan2020exploiting,wang2021sketch,gal2021stylegan}, we explore optimizing only a subset of model weights during training. Surprisingly, we obtain similar performances by tuning either all weights or just the weight of a single layer. Moreover, for single-layer updates, we find that the weight update has a low-rank structure, agreeing with the findings of Bau~\etal~\shortcite{bau2020rewriting}. Taking advantage of the low-rank structure, we can train a model by updating the weight at layer $j$, denoted by $W_j$, where we enforce the weight update to be low-rank as follows:

\begin{equation}
\begin{aligned}
     &\argmin_{U, V} \frac1N \sum_{i=1}^{N} \mathbb{E}_{\lattexture} \left[ \lossproj(\mgspstylayer, \editspstylayer) \right]\\
     &\text{subject to} \;\;\;\;\;\; \;\;\; W_j' = W_j + UV
\end{aligned}
\end{equation}
Following the formulation of \cite{bau2020rewriting}, we view $W_j \in \mathbb{R}^{m\times n}$ as a $m$ by $n$ matrix, where $n$ and $m$ denote the number of input and output channels, respectively for the $1\times 1$ Conv layers used in StyleGAN3~\cite{Karras2021alias}.  We define $U \in \mathbb{R}^{m\times r}$ and $V \in \mathbb{R}^{r\times n}$, where $r$ is the desired rank. As a design choice, we choose a small $r$ such that $r \ll m, n$. Since the optimized parameters $U$ and $V$ are restricted to rank $r$ by their shape, the change in the weight $W_j' - W_j = UV$ has a rank not greater than $r$. The update scheme further reduces the storage requirement for each created model, since we can store just the difference of the weights as $U$ and $V$. At inference time, we can add the difference to the pre-trained model to perform the model edits. 

In our setting, storing the full weight takes around 60 MB, and storing a single layer takes around 4 MB. We observe that $r=50$ performs well, and storing the rank-50 matrices takes around $0.6$MB, reducing the storage by 100$\times$. Efficient storage is critical for real-world applications as we want to avoid storing the weight differences at full scale, each time we create a new effect. %

 We observe that our proposed style-mixing augmentation serves as a strong regularization for the above three variants: full-model updates, single-layer updates, and low-rank updates. We provide a detailed analysis in \refsec{evaluations}.

\begin{figure}
    \centering
    {%
    \setlength{\fboxsep}{0pt}%
    \setlength{\fboxrule}{0.5pt}%
    \fbox{\includegraphics[width=0.9\linewidth]{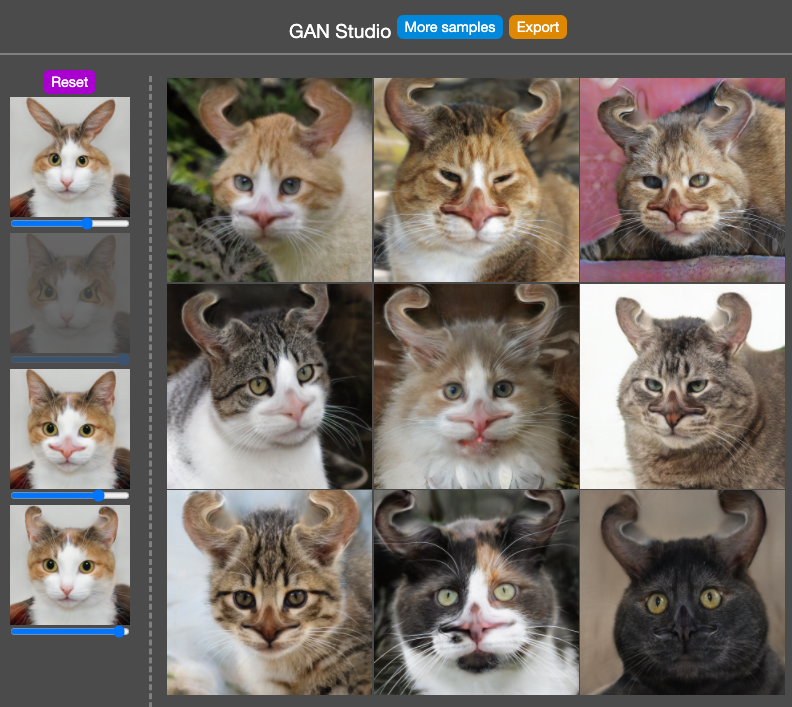}}
    } %
    \caption{{\bf User interface for model composition.} We present an interface for users to easily create a new model by composing the edited models made beforehand. To get the desired model, the user can toggle on and off each edited model on the left and move the slider bars to adjust the blending strength of each model. The user can also click a button to view more samples from the new model and export the created model weights for future applications. Please view the supplemental video for more details.}
    \label{fig:knob_interface}
\end{figure} \subsection{Composing Multiple Edited Models}
\label{sec:compose}
One of our key applications is to enable users to compose different models created beforehand to achieve aggregated effects. The composition operation is powerful since it caters to the group of users who wish to have simple, interactive access to the creation of a new model. For example, users can download models with preset effects shared by other model creators,  and personalize the combination of models.

Surprisingly, the composition can be achieved by a simple linear operation in the weight space. Given a set of models created from the same pre-trained generator, we denote the weights of the pre-trained model as $\theta$ and previously created models as $\{\theta'_1, \theta'_2, \dots, \theta'_K\}$. We can compose a new model of weights $\theta'_{\text{new}}$ as follows:
\begin{equation}
    \theta'_{\text{new}} = \theta + \sum_{k=1}^{K} \alpha_k (\theta'_k - \theta)
\end{equation}
where $\alpha_k$ is a parameter that controls the strength of the edits from each model. 
Based on the finding, we present an interactive user interface to create new models from a set of pre-defined models, as shown in \reffig{knob_interface}. A user can easily control the contribution of each preset to produce aggregated edits within seconds, and export their favorite combinations as a new, customized model.

\subsection{Extending to Color Changes}
\label{sec:color}
Our method can also be applied to color edits. To achieve this, a user is asked to paint certain image regions with selected colors and specify regions where the color needs to be preserved. For each user edit, we have the specified color $C$, the colored region $\mcolor$, and the preserved region $\mfixed$. We use this input to design a loss as follows:

\begin{equation}
\begin{aligned}
     \argmin_{\theta'} \frac1N &\sum_{i=1}^{N} \mathbb{E}_{\lattexture}  \big [ \lcolor ||\mcolor \odot (C - \mgspsty)||_1 \\
      & \;\;\ + ||\mfixed \odot (\ogspsty - \mgspsty)||_1 \big ].
\end{aligned}
\end{equation}

We apply the loss to the masks $\mcolor$ and $\mfixed$, which specify the regions to update or preserve the color,  respectively. We find that our method works well with $\lcolor=8$. We also apply style-mixing augmentation to this task. Since the augmentation does not alter shape, the same color mask can be applied to the augmented samples to achieve similar effects. Our edited models produce realistic, local color changes, thanks to the disentangled representation learned in the source models. For example, we can change the color of the entire sky without precisely painting the whole region. More details are in~\refsec{applications}.

\section{Experiments}

We evaluate our method, along with the baseline methods, on a number of warping instructions in~\refsec{evaluations}. In \refsec{ablation}, we compare different variants of our full method. We also present several applications of our method and the qualitative results in~\refsec{applications}.

\begin{table}[t]
\caption{{\bf Quantitative baseline comparisons.} We compare our rank-50 single layer update method against several baselines. our method outperforms most baselines in both the edited and unaltered regions. \camready{While training with extremely few samples, our method performs on par with CATs~\cite{cho2021cats}, a supervised dense correspondence method.} More details of the baselines and metrics can be found in~\refsec{evaluations}.}
{\small
    \centering
    \resizebox{1.\linewidth}{!}{
\begin{tabular}{clccccccc}
\toprule
\multirow{3}{*}{Class} & \multirow{3}{*}{Name} & \multicolumn{4}{c}{Edited region} & \multicolumn{3}{c}{Unaltered Region} \\ \cmidrule(lr){3-6} \cmidrule(lr){7-9} 
 &  & PSNR & SSIM & LPIPS & CD & PSNR & SSIM & LPIPS \\
 &  & ($\uparrow$) & ($\uparrow$) & ($\downarrow$) & ($\downarrow$) & ($\uparrow$) & ($\uparrow$) & ($\downarrow$) \\ \midrule \midrule
\multirow{10}{*}{Cat} & TGAN & 6.67 & 0.21 & 0.60 & 6.41 & 6.03 & 0.34 & 0.67 \\
 & TGAN+ADA & 6.84 & 0.22 & 0.52 & 7.01 & 7.98 & 0.41 & 0.56 \\
 & FreezeD & 6.34 & 0.20 & 0.58 & 6.92 & 6.47 & 0.34 & 0.64 \\
 & \cite{ojha2021few-shot-gan} & 7.25 & 0.22 & 0.51 & 7.73 & 7.96 & 0.40 & 0.53 \\
 & NADA (text) & 8.67 & 0.24 & 0.62 & 10.98 & 14.32 & 0.50 & 0.28 \\
 & NADA (image) & 8.13 & 0.25 & 0.55 & 9.06 & 14.20 & 0.56 & 0.25 \\
 & Rewriting (rank 1) & 9.73 & 0.26 & 0.53 & 9.75 & 19.91 & 0.76 & 0.10 \\
 & Rewriting (rank 50) & 9.81 & 0.26 & 0.53 & 8.98 & 19.89 & 0.75 & 0.11 \\
 & CATs & \textbf{12.98} & \textbf{0.41} & \textbf{0.24} & \textbf{3.05} & \textbf{23.14} & \textbf{0.83} & \textbf{0.06} \\ \cmidrule(l{5pt}r{5pt}){2-9}
 & Ours & 11.83 & 0.36 & 0.30 & 5.07 & 21.09 & 0.77 & 0.08 \\ \midrule
\multirow{10}{*}{Horse} & TGAN & 8.53 & 0.21 & 0.49 & 8.44 & 8.44 & 0.19 & 0.48 \\
 & TGAN+ADA & 8.15 & 0.19 & 0.48 & 6.23 & 8.43 & 0.18 & 0.46 \\
 & FreezeD & 7.86 & 0.18 & 0.47 & 7.28 & 8.18 & 0.18 & 0.46 \\
 & \cite{ojha2021few-shot-gan} & 8.98 & 0.17 & 0.45 & 8.88 & 8.84 & 0.19 & 0.41 \\
 & NADA (text) & 7.92 & 0.12 & 0.56 & 8.08 & 9.52 & 0.21 & 0.32 \\
 & NADA (image) & 9.23 & 0.14 & 0.54 & 8.07 & 11.05 & 0.27 & 0.25 \\
 & Rewriting (rank 1) & 8.57 & 0.16 & 0.52 & 7.51 & 20.33 & 0.79 & 0.06 \\
 & Rewriting (rank 50) & 8.73 & 0.17 & 0.51 & 9.27 & 19.94 & 0.77 & 0.06 \\
 & CATs & 10.24 & 0.24 & 0.41 & 4.46 & 19.75 & 0.78 & 0.07 \\ \cmidrule(l{5pt}r{5pt}){2-9} 
 & Ours & \textbf{11.98} & \textbf{0.28} & \textbf{0.30} & \textbf{3.87} & \textbf{20.41} & \textbf{0.79} & \textbf{0.05} \\ \midrule
\multirow{10}{*}{House} & TGAN & 5.98 & 0.25 & 0.64 & 9.99 & 7.13 & 0.30 & 0.61 \\
 & TGAN+ADA & 6.64 & 0.28 & 0.58 & 8.05 & 8.07 & 0.32 & 0.56 \\
 & FreezeD & 5.89 & 0.25 & 0.68 & 12.11 & 7.28 & 0.27 & 0.68 \\
 & \cite{ojha2021few-shot-gan} & 6.80 & 0.32 & 0.66 & 13.89 & 8.94 & 0.34 & 0.54 \\
 & NADA (text) & 6.75 & 0.28 & 0.60 & 9.88 & 9.21 & 0.27 & 0.45 \\
 & NADA (image) & 6.89 & 0.31 & 0.64 & 12.17 & 10.53 & 0.34 & 0.38 \\
 & Rewriting (rank 1) & 7.24 & 0.31 & 0.61 & 11.81 & 15.29 & 0.56 & 0.16 \\
 & Rewriting (rank 50) & 7.64 & 0.31 & 0.58 & 10.84 & 14.74 & 0.56 & 0.20 \\
 & CATs & 9.14 & 0.36 & 0.47 & 5.39 & 15.29 & \textbf{0.61} & 0.19 \\ \cmidrule(l{5pt}r{5pt}){2-9} 
 & Ours & \textbf{10.04} & \textbf{0.37} & \textbf{0.37} & \textbf{5.19} & \textbf{16.45} & \textbf{0.60} & \textbf{0.11} \\ \bottomrule
\end{tabular}
}}
\label{tbl:baseline}
\end{table}

\begin{figure}
    \centering
    \includegraphics[width=\linewidth]{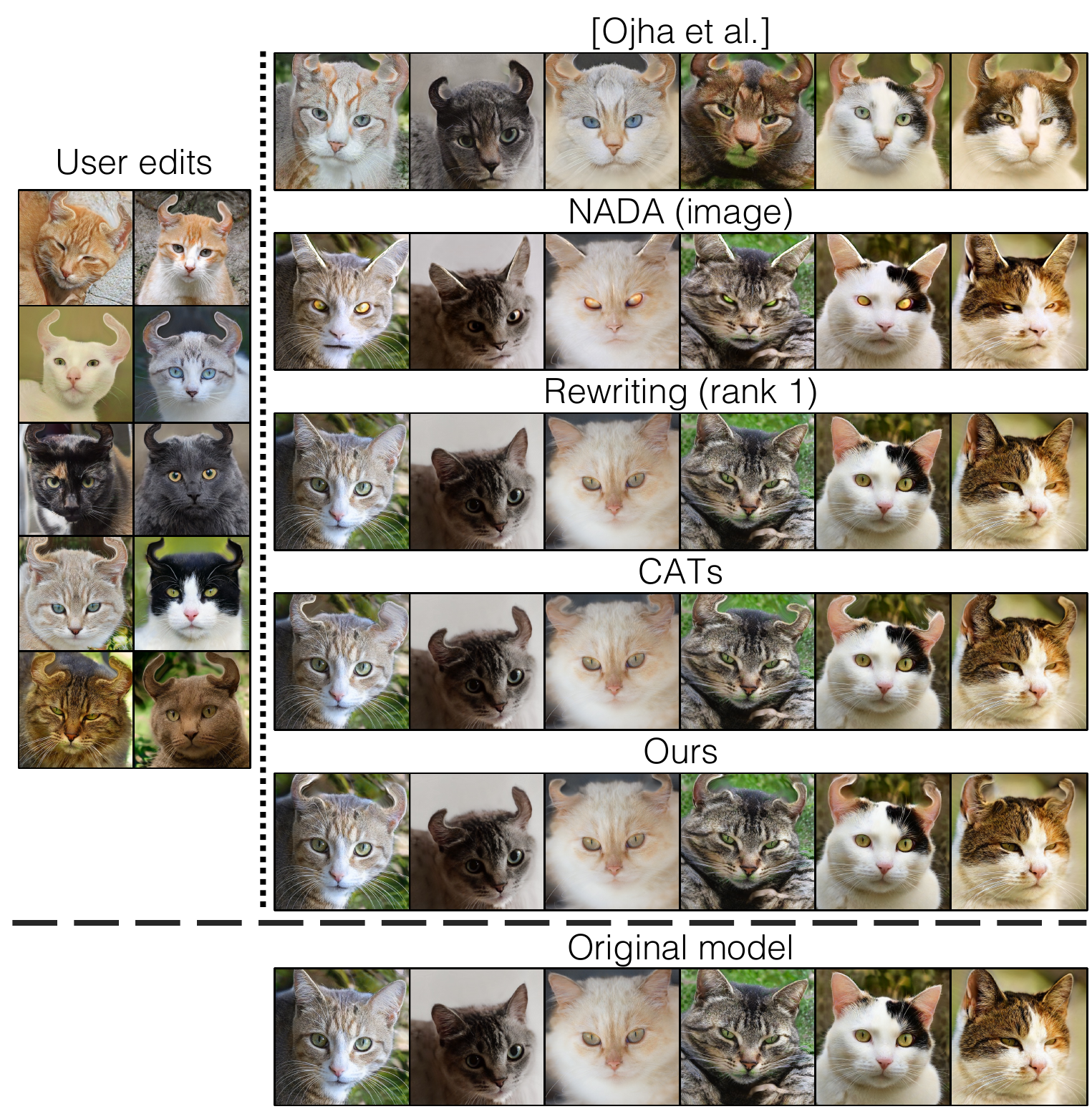}
    \caption{{\bf Qualitative baseline comparisons.} We compare our method against several baselines. For each method, we update the original model (bottom) using 10 user edit examples (left).  We use the same noise $z$ to generate the images. We find that our method creates a model that matches the specified shape changes, while preserving the color and texture of cat faces and backgrounds. On the other hand, the few-shot GAN fine-tuning method by Ojha~\etal~\shortcite{ojha2021few-shot-gan} alters the color and textures, and both StyleGAN-NADA~\cite{gal2021stylegan} and Model Rewriting~\cite{bau2020rewriting} fails to apply accurate shape changes. \camready{Applying CATs, a supervised dense correspondence model, can achieve similar results but with much higher inference costs in both run-time and memory. More details can be found in~\refsec{evaluations}.}}
    \label{fig:baseline}
\end{figure}
\begin{figure}
    \centering
    \includegraphics[width=\linewidth]{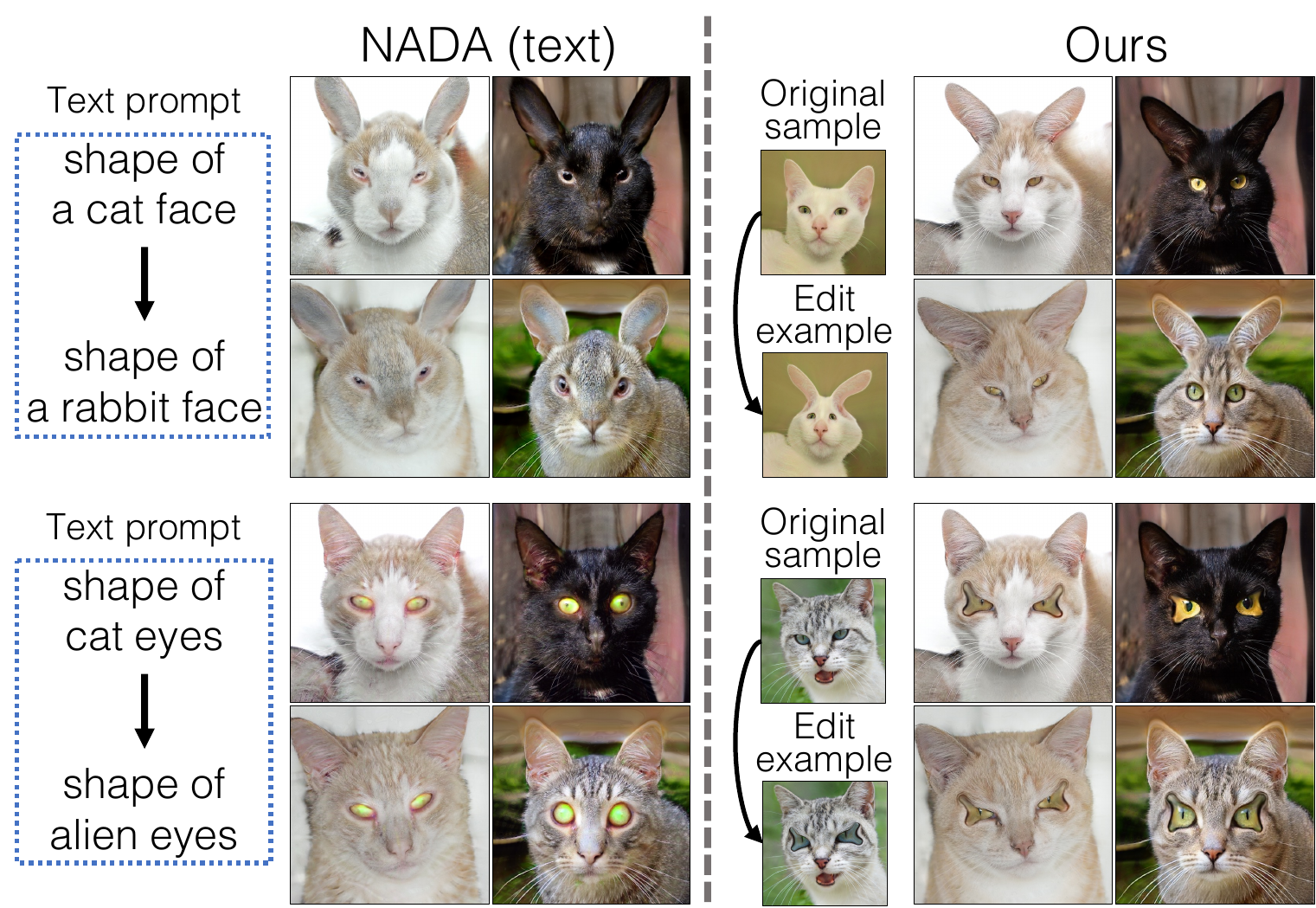}
    \caption{{\bf Comparison with text-based model editing.}  We test if the text-based model editing proposed by StyleGAN-NADA~\cite{gal2021stylegan} can achieve the warping edits. It is unnatural to describe the warping edits precisely using text, but we attempted to provide the text prompts (left) that best match the warping edits (right). Despite this, we observe that training with text inputs leads to unintended color and texture changes.}
    \label{fig:nada_comparison}
\end{figure}

\paragraph{Implementation details.} 
We adopt StyleGAN3-R~\cite{Karras2021alias}, as the model learns better geometric representations along with translation and rotation invariance. The style-mixing operation is performed in the intermediate latent space produced by the mapping network. Also, we fix the weights for the mapping network for full-weight updates. %

We train and test our models on a single NVIDIA A5000 GPU. We train all models with 2000 iterations using an Adam optimizer~\cite{kingma2015adam}. We adopt the learning rate schedule from the image projection method proposed by Karras et al.~\shortcite{karras2020analyzing}, where the learning rate is ramped up from zero linearly during the first 100 iterations and ramped down to zero using a cosine schedule during the last 1000 iterations. We use a learning rate of 0.05 and a batch size of 8 for single-layer updates. For full-model updates,  we use a learning rate of 0.001 and a batch size of 6. It takes  20 minutes to train a model of $256\times 256$ resolution and 40 minutes to train a model of $512\times 512$ resolution. We find that updating either the full weights or a single layer results in a similar training time.

\subsection{Evaluations}
\label{sec:evaluations}
\paragraph{Test cases.}
We demonstrate our model editing method in various ways. 
We evaluate our method on three source models, which are pre-trained on AFHQv2 cats~\cite{choi2020starganv2}, LSUN horses~\cite{yu2015lsun}, and Places houses~\cite{zhou2017places}. We perform four different edit operations on the cat model and two on the horse and house model. For every operation, we create ten edit examples for training and five hold-out edit examples for validation and testing. Additionally, we would like to evaluate the edited and unaltered regions separately. To achieve this, we manually mask the edited regions in the edit examples and only use the masks for validation and testing. %

\paragraph{Performance metrics.}
We evaluate our method in two ways. First, we expect a decent model to successfully transfer user edits to other unseen samples, wherever the edits are applicable. Also, we hope the model to preserve the content of unaltered regions. In particular, given the edit example and the corresponding output from the edited model, we apply similarity measures separately to the edited region and the untouched region as defined above. The similarity is measured by three standard metrics: Peak Signal-to-Noise Ratio (PSNR), Structural Similarity Index (SSIM)~\cite{wang2004image}, and LPIPS~\cite{zhang2018unreasonable}.

To directly assess the correctness of shape changes, we adopt the procedure from Wang et al.~\shortcite{wang2021sketch}, where the symmetric Chamfer Distance (CD)~\cite{barrow1977parametric} $d(x, y) + d(y, x)$ is calculated between the contour of the two images $x, y$. The contours are computed by DexiNed~\cite{xsoria2020dexined} and further processed to 1-pixel-wide edges as described in Isola et al.~\shortcite{isola2017image}. We report the Chamfer Distance in the edited region.

We average the numbers over all the test cases from the same source model and report the results for each object category. %

\paragraph{Baseline comparisons.}
We compare our method against several recent few-shot GAN fine-tuning methods: TGAN~\cite{wang2018transferring}, TGAN+ADA \cite{karras2020ADA}, FreezeD~\cite{mo2020freeze}, Ojha et al.~\shortcite{ojha2021few-shot-gan}, and StyleGAN-NADA~\cite{gal2021stylegan}. The first four methods are directly trained on the edit examples, and StyleGAN-NADA is trained on CLIP embeddings~\cite{radford2021learning}. The CLIP embeddings are generated either from text inputs or image inputs, and we denote the two baselines by \emph{NADA (text)} and \emph{NADA (image)}, respectively. We acknowledge that it is unnatural to describe the warping edits precisely using text. However, to demonstrate the inherent differences between the use of text prompts and warping interfaces in our task, we have attempted to provide text prompts that best match the desired warping effect. For \emph{NADA (image)}, we extract the target CLIP embeddings from the same edit examples. 

\camready{In addition, we compare our work with Model Rewriting~\cite{bau2020rewriting}, a method that applies low-rank updates to a model layer for copy-and-paste edits. We apply Model Rewriting to our warping tasks with rank-1 and rank-50 updates. Also, we compare against a baseline based on the state-of-the-art supervised dense correspondence algorithm CATs~\cite{cho2021cats}. We estimate the correspondences between the edit examples and every unseen sample, which we use to transfer the user edits. More implementation details of baselines are in the \suppmat{supplemental material}.}

\reftbl{baseline} shows the quantitative comparisons. We find that our method outperforms few-shot GANs, StyleGAN-NADA, and Model Rewriting by far, and this result agrees with the qualitative comparisons presented in \reffig{baseline}. Few-shot GAN fine-tuning methods, such as the state-of-the-art by Ojha et al.~\shortcite{ojha2021few-shot-gan}, are not designed to preserve the color and texture of the original model. Instead, these methods alter the textures to match a small number of training samples. \camready{Notably, our problem setting differs from few-shot adaptation in two ways. First, we aim to enable a user to directly control the behavior of a pre-trained model, instead of adapting a model to a new domain, Second, our method learns to mimic before-and-after user edits, instead of matching the output distribution.} 

NADA (image) preserves the color and textures better but fails to apply accurate warping edits. We speculate that CLIP embeddings may not capture sufficient fine-grained information of the image, or may not generalize to out-of-distribution shapes. We also provide a qualitative comparison against NADA (text) in \reffig{nada_comparison}. Despite our effort to explicitly specify shape via text alone, training with text inputs leads to unwanted color and texture changes. %
\camready{We find that Model Rewriting does not change the object shapes, since the method's training objective is specialized for copy-and-paste edits. On the other hand, the CATs baseline works on par with ours. Trained with dense correspondence supervision, this baseline outperforms our method for cat models, whereas our methods works better for horse and house models. We note that our method is only trained with extremely few samples and takes only one forward pass to create new samples. In contrast, CATs-based warping transfer requires us to align a new sample to all the existing training samples via the dense correspondence algorithm.  Also,  directly transferring warping fields does not accommodate dramatic changes in object sizes,  %
which hurts the baseline's performance for models with large geometric variations (e.g., horse and house model).}

\begin{table}[t]
\caption{{\bf Ablation studies.} We evaluate the effects of style-mixing augmentation \textbf{(style-mix aug.)} and the differences between full-weight updates \textbf{(Full)}, single-layer updates \textbf{(Layer)}, and single-layer updates with a rank-50 threshold \textbf{(Layer (r50))}. Note that style-mixing augmentation consistently improves all cases. The full-weight update method achieves the best performance, while single-layer methods obtain comparable performance with less demanding storage requirements.}
{\small
    \centering
    \resizebox{1.\linewidth}{!}{
\begin{tabular}{clccccccc}
\toprule
\multirow{3}{*}{Class} & \multirow{3}{*}{Name} & \multicolumn{4}{c}{Edited region} & \multicolumn{3}{c}{Unaltered Region} \\  \cmidrule(lr){3-6} \cmidrule(lr){7-9} 
 &  & PSNR & SSIM & LPIPS & CD & PSNR & SSIM  & LPIPS \\
 & & ($\uparrow$) & ($\uparrow$) & ($\downarrow$) & ($\downarrow$) & ($\uparrow$) & ($\uparrow$) & ($\downarrow$) \\ \midrule \midrule

\multirow{6}{*}{Cat} & Full & 11.60 & 0.37 & 0.40 & 7.61 & 21.89 & \textbf{0.80} & 0.08 \\
 & Full + style-mix aug. & \textbf{12.56} & \textbf{0.40} & \textbf{0.28} & \textbf{4.49} & \textbf{22.07} & \textbf{0.80} & \textbf{0.07} \\
 & Layer & 11.56 & 0.35 & 0.39 & 6.40 & 19.46 & 0.73 & 0.11 \\
 & Layer + style-mix aug. & 11.90 & 0.36 & 0.30 & 5.18 & 21.04 & 0.77 & 0.08 \\
 & Layer (r50) & 11.50 & 0.36 & 0.35 & 6.18 & 19.01 & 0.70 & 0.12 \\
 & Layer (r50) + style-mix aug. & 11.83 & 0.36 & 0.30 & 5.07 & 21.09 & 0.77 & 0.08 \\ \midrule

\multirow{6}{*}{Horse} & Full & 10.09 & 0.26 & 0.43 & 5.28 & 19.45 & 0.74 & 0.08 \\
 & Full + style-mix aug. & \textbf{12.55} & \textbf{0.31} & \textbf{0.25} & \textbf{3.23} & 19.05 & 0.71 & 0.06 \\
 & Layer & 10.33 & 0.26 & 0.39 & 4.09 & 18.09 & 0.69 & 0.10 \\
 & Layer + style-mix aug. & 11.81 & 0.27 & 0.31 & 4.06 & 19.96 & 0.77 & 0.06 \\
 & Layer (r50) & 10.36 & 0.26 & 0.37 & 4.86 & 16.99 & 0.63 & 0.12 \\
 & Layer (r50) + style-mix aug. & 11.98 & 0.28 & 0.30 & 3.87 & \textbf{20.41} & \textbf{0.79} & \textbf{0.05} \\ \midrule

\multirow{6}{*}{House} & Full & 8.99 & 0.37 & 0.50 & 6.71 & 17.23 & \textbf{0.64} & 0.15 \\
 & Full + style-mix aug. & \textbf{10.55} & \textbf{0.39} & \textbf{0.36} & \textbf{4.63} & \textbf{17.27} & \textbf{0.64} & \textbf{0.11} \\
 & Layer & 9.64 & 0.37 & 0.41 & 5.08 & 16.01 & 0.59 & 0.13 \\
 & Layer + style-mix aug. & 10.00 & 0.37 & 0.37 & 5.16 & 16.51 & 0.60 & \textbf{0.11} \\
 & Layer (r50) & 10.04 & 0.37 & 0.39 & 5.13 & 15.40 & 0.56 & 0.14 \\
 & Layer (r50) + style-mix aug. & 10.04 & 0.37 & 0.37 & 5.19 & 16.45 & 0.60 & \textbf{0.11} \\ \bottomrule
\end{tabular}
}}
\label{tbl:ablation}
\end{table}
\begin{figure}
    \centering
    \includegraphics[width=\linewidth]{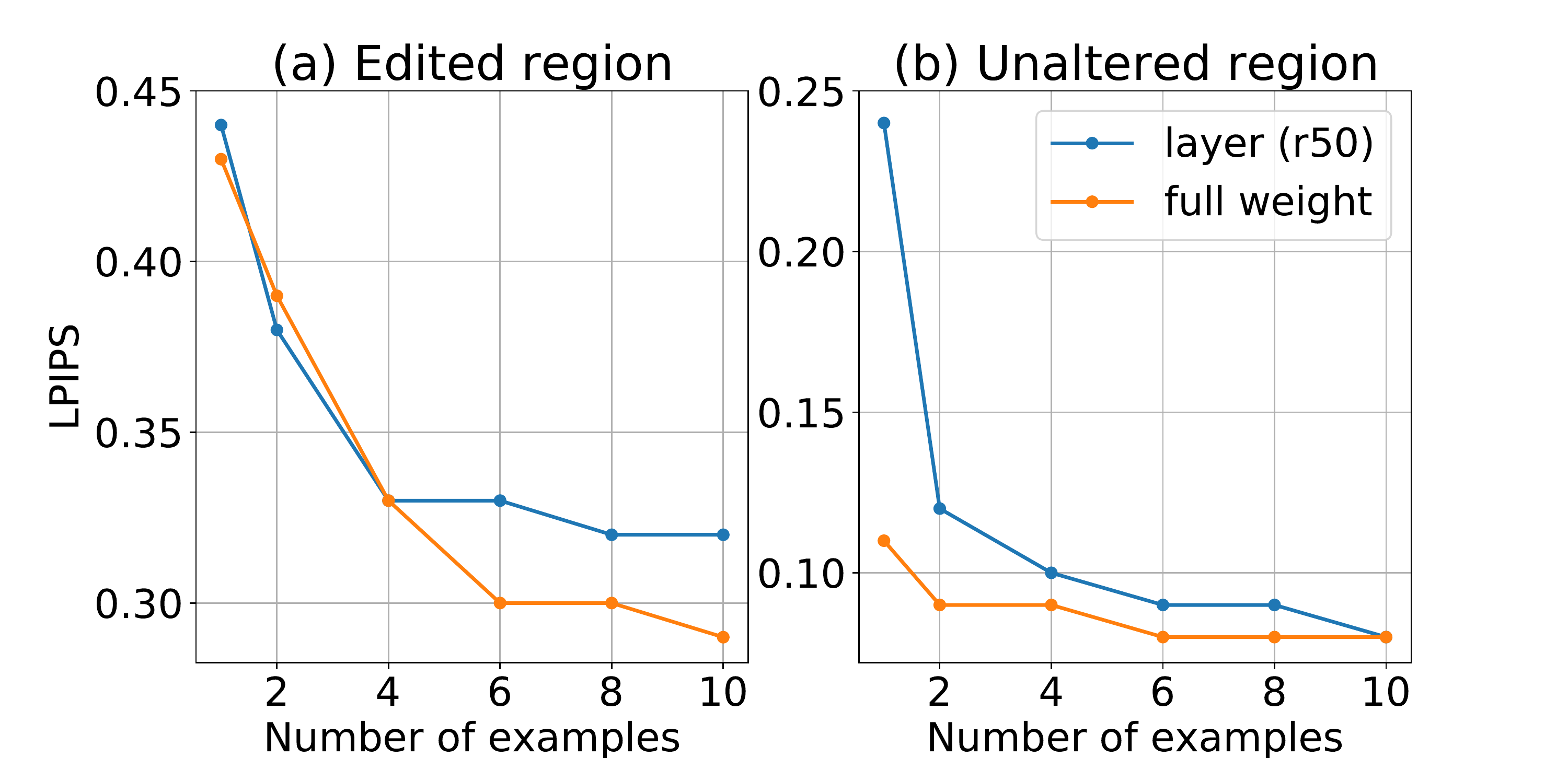}
    \caption{{\bf Training with fewer user edit examples.} We study how the number of user edit examples affects our model's performance. For both full-weight updates and rank 50 single-layer updates (``layer (r50)''), the performance improves when more examples are added, and the improvement quickly saturates after training with four or more examples. Both models are trained with style-mixing augmentation, and we report average LPIPS over all test cases.}
    \label{fig:num_sample}
\end{figure}

\begin{figure}
    \centering
    \includegraphics[width=\linewidth]{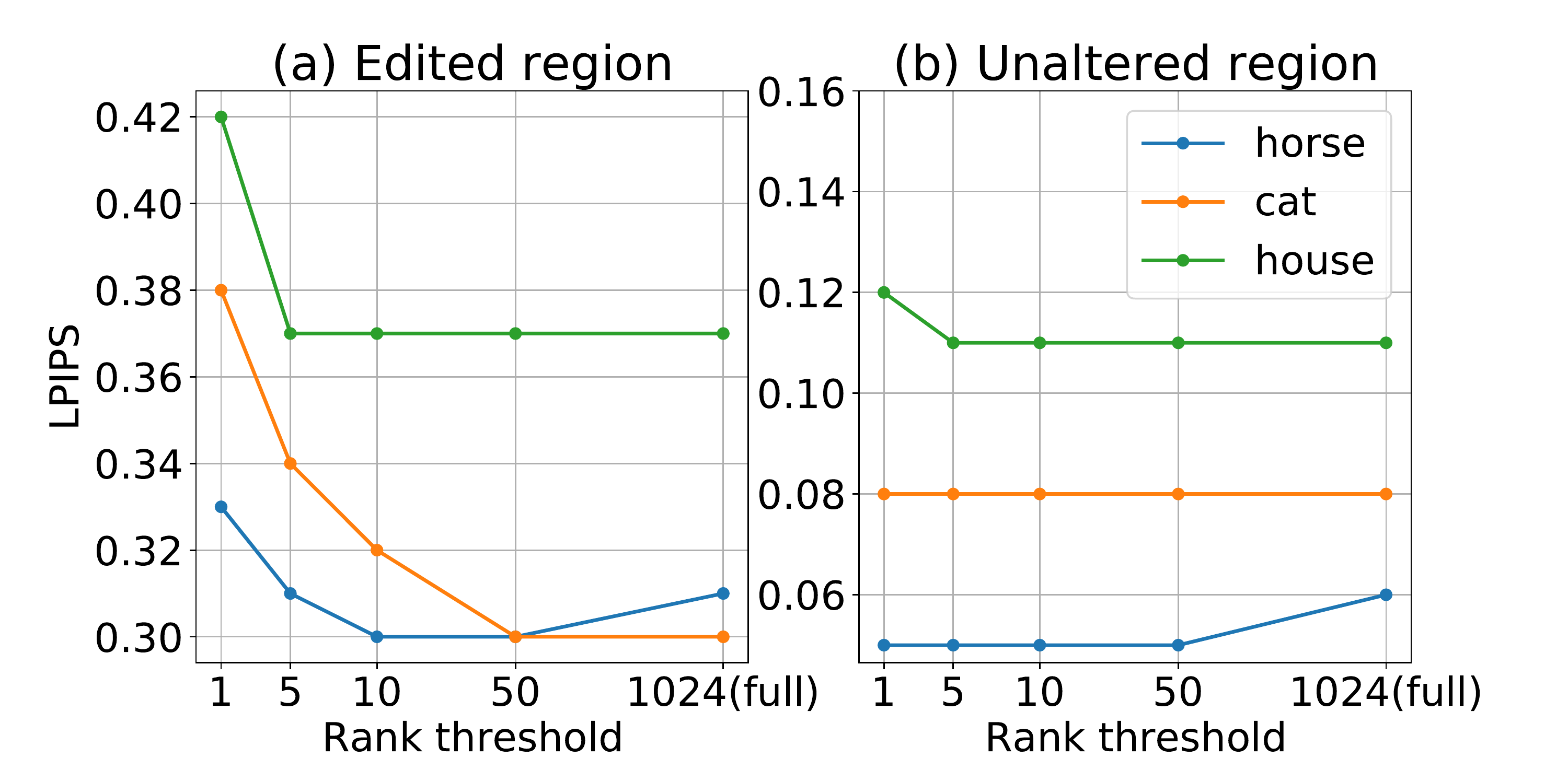}
    \caption{{\bf Quantitative analysis on different rank thresholds.} We analyze the low-rank structure of our single-layer update methods by evaluating models trained on different rank thresholds. In all classes of models, we observe that the performance saturates quickly as the rank threshold increases. All models are trained with style-mixing augmentation, and we report performance with LPIPS averaged over each class.}
    \label{fig:num_rank}
\end{figure}
\begin{figure}
    \centering
    \includegraphics[width=\linewidth]{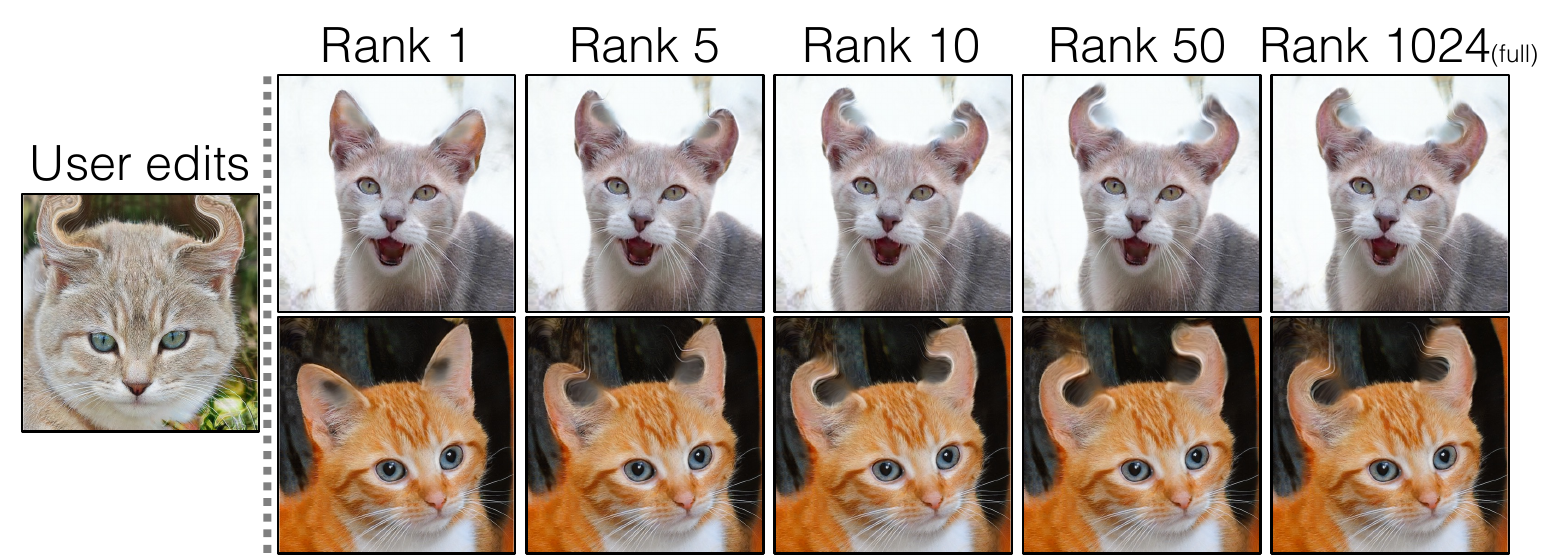}
    \caption{{\bf Qualitative analysis on different rank thresholds.} We visually compare the models trained with different rank thresholds. When the rank threshold is too small (rank 1, 5), the model fails to produce sharp changes to the cat ears. However, the results improve quickly as the rank threshold increases, and a desirable effect can be achieved with a rank-50 update.}
    \label{fig:rank_qual}
\end{figure}

\subsection{Ablation Study}
\label{sec:ablation}
We investigate our method's performance regarding several algorithmic choices and hyperparameters.  The quantitative analysis is summarized in \reftbl{ablation}.

\begin{figure*}
    \centering
    \includegraphics[width=\linewidth]{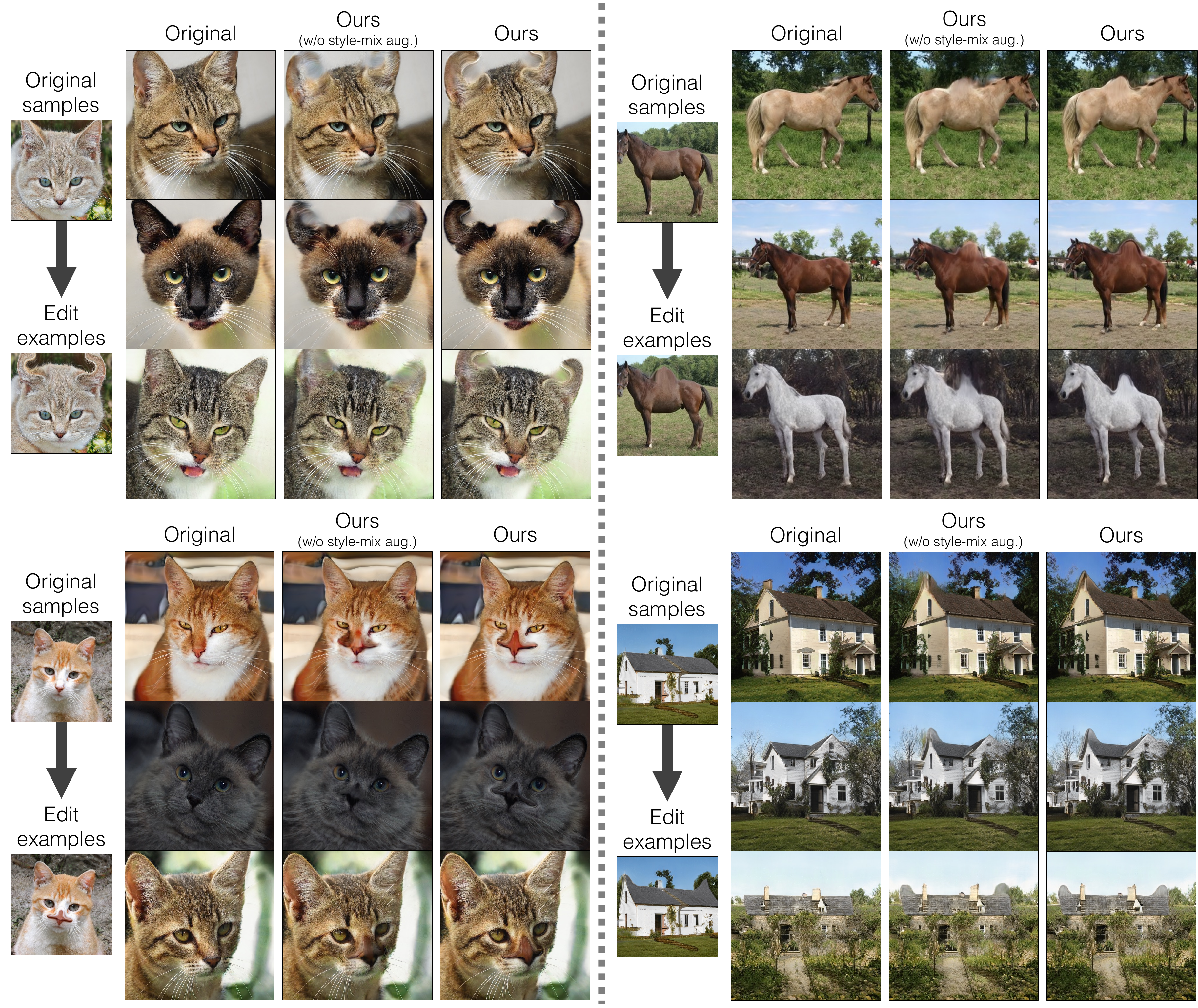}
    \caption{{\bf Style-mixing augmentation improves generalization.} We show the effects of style-mixing augmentation in 4 editing tasks. Each block shows one model editing task. For each task, 10 user edit examples are used for training, and one example is shown. We show samples generated from the original models and the edited models trained with or without style-mixing augmentation. The edited models are all trained with the rank-50, single-layer update method, as explained in~\refsec{parameter}. For each task, we show samples generated from the same three $z$. We note that models trained without augmentation often generate blurry artifacts in the edited region. In contrast, the models trained with augmentation consistently generate crisper shape changes. The readers are encouraged to zoom in and view the results in more detail.}
    \label{fig:stylemix_effect}
\end{figure*}

\paragraph{Style-mixing augmentation.} 
Applying style-mixing augmentation improves performance consistently in all cases, showing that the augmentation largely increases model generalization. %
This finding is also consistent with the qualitative results as shown in \reffig{stylemix_effect}, where the augmented models generate more promising ears and noses shape changes in the cat model and produce crisper and more accurate edits on the horse backs and house roofs. As such, the augmented models obtain better similarity scores and Chamfer Distances in the edited region. \reftbl{ablation} also shows that augmented models barely alter the untouched region, since they are trained to preserve samples with a wide range of color and texture variations.

\paragraph{Updating fewer parameters.}  
We test the effects of updating the full model, a single layer, or a single layer with rank-50 update as described in~\refsec{parameter}. \reftbl{ablation} shows that the performances of the three strategies are similar, with the best-performing models being tuned with full weights. Interestingly, models updated with full weights improve more than other models in the edited regions after applying style-mixing augmentation. 
This suggests that style-mixing augmentation serves as a strong regularization to prevent model overfitting even when updating the full weights, and it avoids the need to update fewer parameters, which is vital for generalization as suggested by previous works~\cite{bau2020rewriting,gal2021stylegan}. Nonetheless, models updated with fewer parameters enjoy smaller storage of weights, and with a similar performance, smaller storage is favorable when a user creates and stores many models. Hence, we generate qualitative results from the models with low-rank updates. %

\paragraph{Which layer to update.}
For single-layer update methods, we select the optimal layer to tune by empirically evaluating on the validation set. We observe that there is a consistent pattern for the optimal layers. In all three source models we tested, we can tune the 2nd convolution layer for coarse geometry changes and the 9th layer for finer, part-based deformations. %

Our method offers options for coarse or fine shape changes, which correspond to updating the 2nd or 9th layer, respectively. The user can select the option that achieves the most desirable effect.

\paragraph{Fewer user edits for training.} %
Next, we investigate how many user edits are sufficient for each weight update method, as we would like to collect as few user edits as possible. As shown in~\reffig{num_sample}, for both full-weight updates and low-rank updates trained with style-mixing augmentation, the performance improves when more examples are added, and the improvement quickly saturates after training with 4 or more examples. This shows that style-mixing augmentation is effective with fewer training samples as well.  %
As a result, we can achieve similar performance with just 4 to 6 user edit examples per warping task.

\paragraph{Different rank thresholds.}
We experimented with different rank thresholds for the low-rank update method, and the results are shown in \reffig{num_rank}. Applying different rank thresholds mainly affects the edited region, where the performance improves but saturates quickly as the rank threshold increases. This observation shows that we can model the warping edits with a small but sufficient number of parameters. Also, the performance saturates at a different rate for different object categories, but we can pick the rank threshold to be 50 for all models to achieve a good balance. Interestingly,  updating with full-rank can sometimes lead to slightly worse performance, as shown in the horse category results. We note that updating extra parameters in a single layer sometimes leads to undesirable changes.

We demonstrate the visual effects of different rank thresholds in \reffig{rank_qual}. When the rank threshold is too low, the model cannot model the geometric changes precisely. But the model can quickly achieve the desirable edit effects as the rank threshold increases.

\begin{table}[t]
\caption{\camready{{\bf Sensitivity to user edit variance.} For every task, we manually create another set of edits, producing a similar geometry change on the same training set. We train 10 models on different combinations of user edits by randomly sampling which edit version to use for each sample. We  report mean $\pm$ stdev over the 10 models. We find that the performance is not sensitive to the user edit variance.}}
{\small
    \centering
    \resizebox{1.\linewidth}{!}{
\begin{tabular}{cccccccc}
\toprule
\multirow{2}{*}{Class} & \multicolumn{4}{c}{Edited region} & \multicolumn{3}{c}{Unaltered region} \\ \cmidrule(lr){2-5} \cmidrule(lr){6-8} 
 & PSNR & SSIM & LPIPS & Chamfer & PSNR & SSIM & LPIPS \\ \midrule \midrule
Cat & 11.83$\pm$0.16 & 0.36$\pm$0.01 & 0.31$\pm$0.01 & 5.19$\pm$0.30 & 20.91$\pm$0.15 & 0.77$\pm$0.00 & 0.08$\pm$0.00 \\
Horse & 11.90$\pm$0.15 & 0.28$\pm$0.01 & 0.30$\pm$0.01 & 3.99$\pm$0.43 & 20.62$\pm$0.10 & 0.79$\pm$0.00 & 0.05$\pm$0.00 \\
House & 10.17$\pm$0.09 & 0.37$\pm$0.00 & 0.36$\pm$0.01 & 4.93$\pm$0.26 & 16.37$\pm$0.09 & 0.60$\pm$0.00 & 0.11$\pm$0.00 \\ \bottomrule
\end{tabular}
}}
\label{tbl:user_noise}
\end{table} %
\begin{table}[t]
\caption{\camready{{\bf Sensitivity to random sample selections.} For every task, we evaluate sensitivity to sample selection on 10 models, where each is trained on a different random subset of $N$ samples. We report  mean $\pm$ stdev over 10 models. We find that the performance does not vary much with different random samples.}}
{\small
    \centering
    \resizebox{1.\linewidth}{!}{
\begin{tabular}{ccccccccc}
\toprule
\multirow{2}{*}{Class} & \multirow{2}{*}{$N$} & \multicolumn{4}{c}{Edited region} & \multicolumn{3}{c}{Unaltered region} \\ \cmidrule(lr){3-6} \cmidrule(lr){7-9} 
 &  & PSNR & SSIM & LPIPS & Chamfer & PSNR & SSIM & LPIPS \\ \midrule \midrule
\multirow{2}{*}{Cat} & 5 & 11.71$\pm$0.23 & 0.36$\pm$0.01 & 0.31$\pm$0.01 & 5.22$\pm$0.40 & 20.55$\pm$0.35 & 0.75$\pm$0.01 & 0.09$\pm$0.01 \\
 & 8 & 11.82$\pm$0.15 & 0.36$\pm$0.01 & 0.31$\pm$0.01 & 5.29$\pm$0.34 & 20.87$\pm$0.17 & 0.77$\pm$0.00 & 0.09$\pm$0.00 \\ \midrule
\multirow{2}{*}{Horse} & 5 & 11.75$\pm$0.43 & 0.28$\pm$0.02 & 0.31$\pm$0.02 & 3.75$\pm$0.80 & 19.92$\pm$0.35 & 0.77$\pm$0.02 & 0.06$\pm$0.00 \\
 & 8 & 11.95$\pm$0.28 & 0.28$\pm$0.01 & 0.30$\pm$0.01 & 3.91$\pm$0.79 & 20.26$\pm$0.18 & 0.78$\pm$0.01 & 0.06$\pm$0.00 \\ \midrule
\multirow{2}{*}{House} & 5 & 9.65$\pm$0.36 & 0.35$\pm$0.01 & 0.39$\pm$0.02 & 5.43$\pm$0.81 & 16.06$\pm$0.30 & 0.59$\pm$0.01 & 0.13$\pm$0.01 \\
 & 8 & 9.89$\pm$0.12 & 0.36$\pm$0.01 & 0.38$\pm$0.01 & 5.14$\pm$0.39 & 16.40$\pm$0.18 & 0.61$\pm$0.01 & 0.12$\pm$0.00 \\ \bottomrule
\end{tabular}
}}
\label{tbl:shuffle}
\end{table} %

\paragraph{Sensitivity to user edit variance.}
\camready{We study how much performance varies if the same user performs editing multiple times. For each task, we evaluate sensitivity by manually creating another set of edits, producing a similar geometry change on the same training set. Models trained on the new edits obtain similar performance to the original ones. Our edited cat models scored 11.83 and 21.09 in PSNR for edited and unaltered regions, respectively; training on the replicated edits gives a similar score (11.83, 20.75). Furthermore, as shown in~\reftbl{user_noise}, we train 10 models on different combinations of user edits by randomly sampling which edit version to use for each sample, and results are consistent and robust: PSNR of the 10 models’ performances is 11.83+/-0.16 and 20.9+/-0.15 (mean+/-stdev) for edited and unaltered regions, respectively.}

\paragraph{Sensitivity to random sample selections.}
\camready{We study our model's sensitivity to different random samples. For each task, we train 10 models to produce the same edits, where each is trained on a different random subset of N samples. As shown in~\reftbl{shuffle}, the variance is small. For cat models, N=5 yields 11.71+/-0.23 and 20.55+/-0.35 (mean+/-stdev) in PSNR for edited and unaltered regions, respectively. N=8 yields 11.82+/-0.15 and 20.87+/-0.17.}

\begin{figure}
    \centering
    \includegraphics[width=\linewidth]{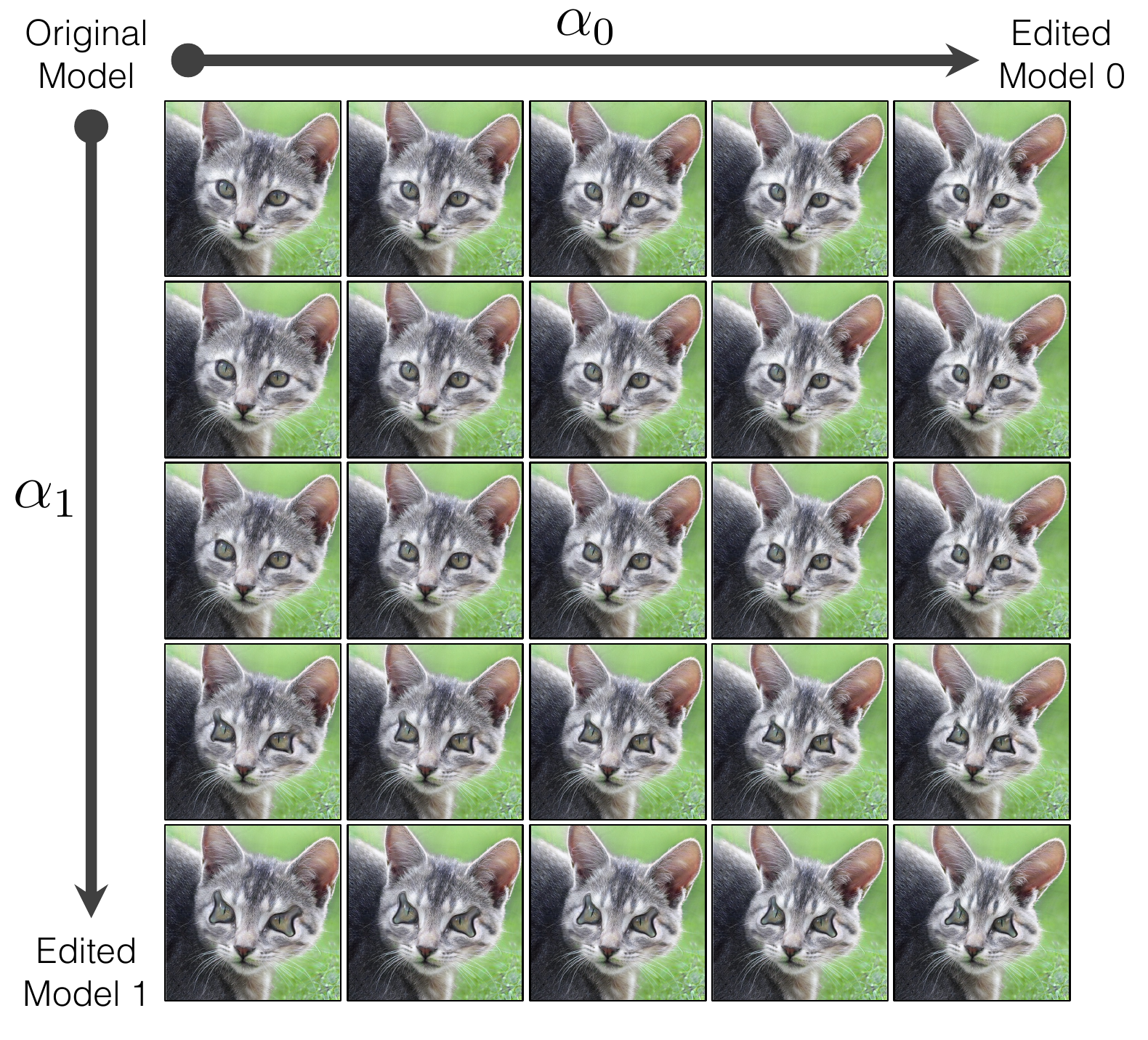}
    \caption{{\bf Composing and interpolating two edited models.} We compose two edited models: one model changes the face shape (top right), and the other changes the eye shape (bottom left). The combined model (bottom right) changes the shape of both faces and eyes. One can adjust the blending strength of each model ($\alpha_0, \alpha_1$) to acquire a smooth shape transition.}
    \label{fig:model_interp}
\end{figure}

\subsection{Applications}
\label{sec:applications}

In this section, we discuss several image editing and synthesis applications based on our method. We show that a user can compose multiple models into a new model, generate smooth transitions between images generated by our models, manipulate real photos, and edit models with color strokes.

\paragraph{Compose edited models.}
We can compose different models to achieve aggregated edit effects, based on the approach explained in~\refsec{compose}. \reffig{model_interp} shows the result of interpolating and composing two models by linearly combining the weights. It is to our surprise that we can achieve smooth shape transitions, even though the newly introduced geometry is not within the original distribution of the source model. Moreover, by applying simple arithmetic to the parameters, we can naturally compose two different out-of-distribution shape changes  and synthesize high-quality samples.

Gal et al.~\shortcite{gal2021stylegan} and Wang et al.~\shortcite{wang2021sketch} have also observed smooth transitions by interpolating the weights between generative models trained by their method. Our results concur with their findings and further suggest the potential to compose out-of-distribution geometric changes for different models.

\begin{figure}
    \centering
    \includegraphics[width=\linewidth]{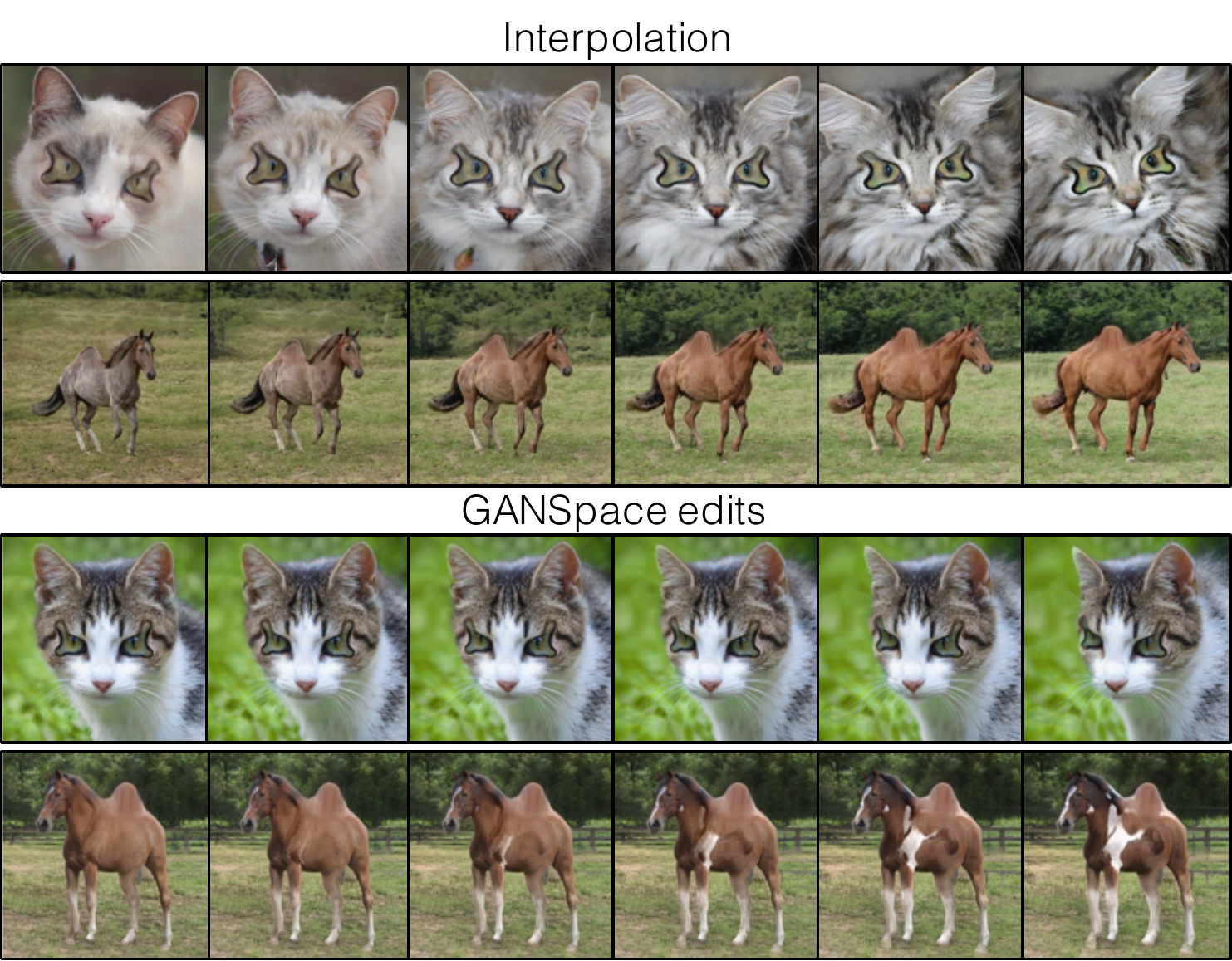}
    \caption{{\bf Smooth latent traversal.} Our edited models can generate smooth transitions between two random samples by interpolating the latent space. We can also apply GANSpace edits~\cite{harkonen2020ganspace} to our models to change the object attributes such as poses or colors.}
    \label{fig:smooth_latent}
\end{figure}

\paragraph{Smooth latent traversals.}
Various useful image manipulations can be done with GANs thanks to its smooth latent space representation. As shown in~\reffig{smooth_latent}, our edited models preserve such properties. First, the edited models can generate smooth transitions between two images by interpolating two random latent codes. Moreover, we can apply interpretable controls to the generated samples. In particular, we obtain the controls from the pre-trained source models using the latent discovery method GANSpace~\cite{harkonen2020ganspace}. The same controls can be applied to the edited models to achieve the same effects such as changing poses or colors. The results indicate that our edited models preserve many desirable properties in the original latent space.%

\begin{figure}
    \centering
    \includegraphics[width=\linewidth]{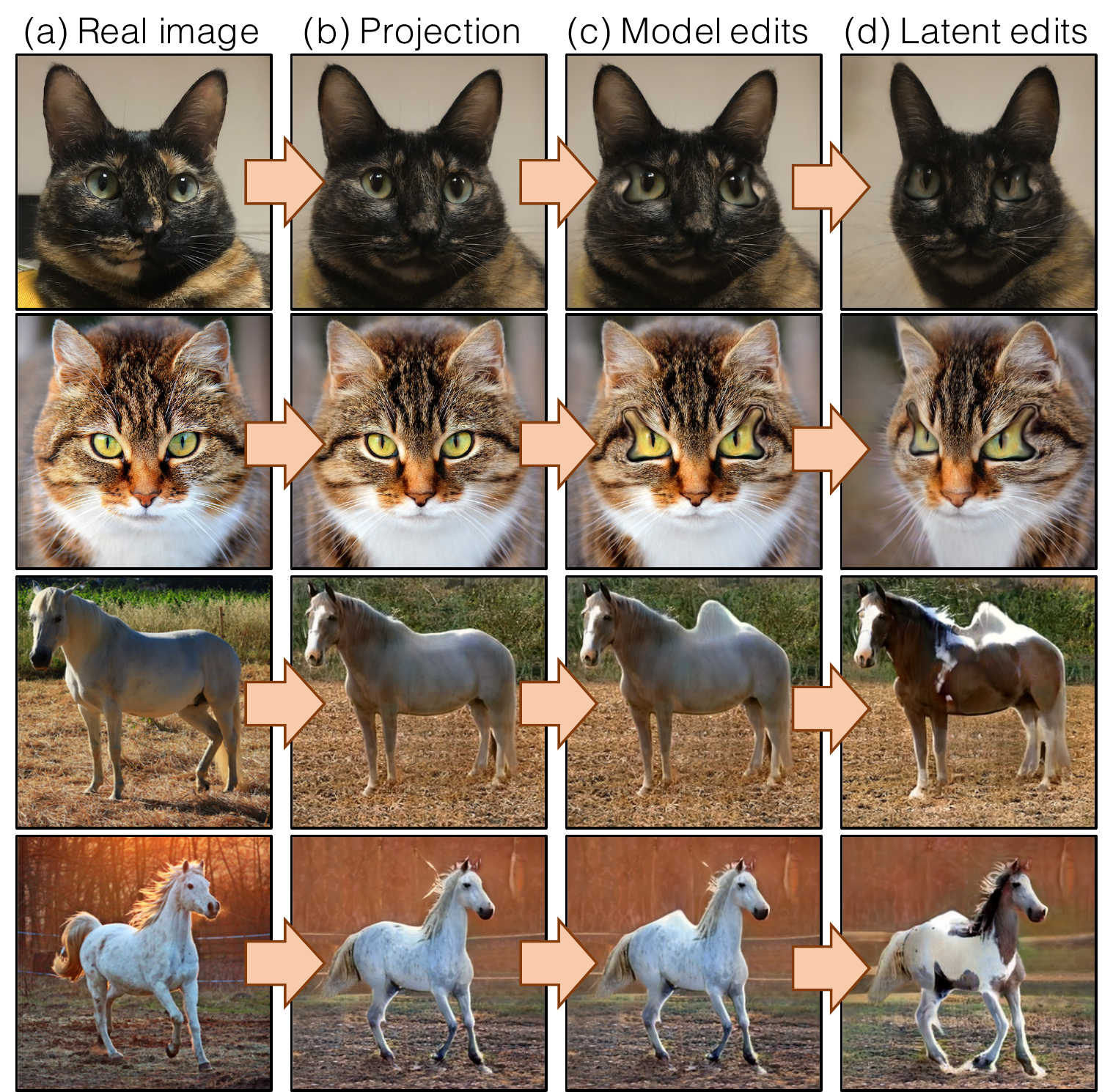}
    \caption{{\bf Editing a real photo.} (a) Given a real photo as input, we can edit it in the following steps: (b) we project the image to the latent space of the source model. (c) We feed the projected latent code to an edited model to effectively transfer the warping edits to the photo. (d) We can further apply the GANSpace edits shown in~\reffig{smooth_latent} to further manipulate the photo. \camready{Real images in the 2nd, 3rd and 4th row courtesy of Pixabay users ``Ben\_Kerckx'', ``rihaij'', and ``rauschenberger'', respectively.} }
    \label{fig:real_image_edit}
\end{figure}

 \paragraph{Real photo manipulation}
We can apply our edited models to manipulate a real photo from the source domain. Specifically, we can transfer the warping edits to the photo and further apply latent manipulations, as shown in~\reffig{real_image_edit}. To achieve this, we project a real photo to a latent code in the source model using StyleGAN2 projection code~\cite{karras2020analyzing}. We feed the projected latent code to the edited model, which effectively transfers the warping effects to the photo. In addition, we can apply latent manipulations to further edit other aspects of the transformed photo. 

In~\reffig{real_image_edit}, we project photos of cat faces into the $W+$ space, an extended latent space introduced by Image2StyleGAN~\cite{abdal2020image2stylegan}. For horses, we project photos into the $W$ space. %

\begin{figure}
    \centering
    \includegraphics[width=\linewidth]{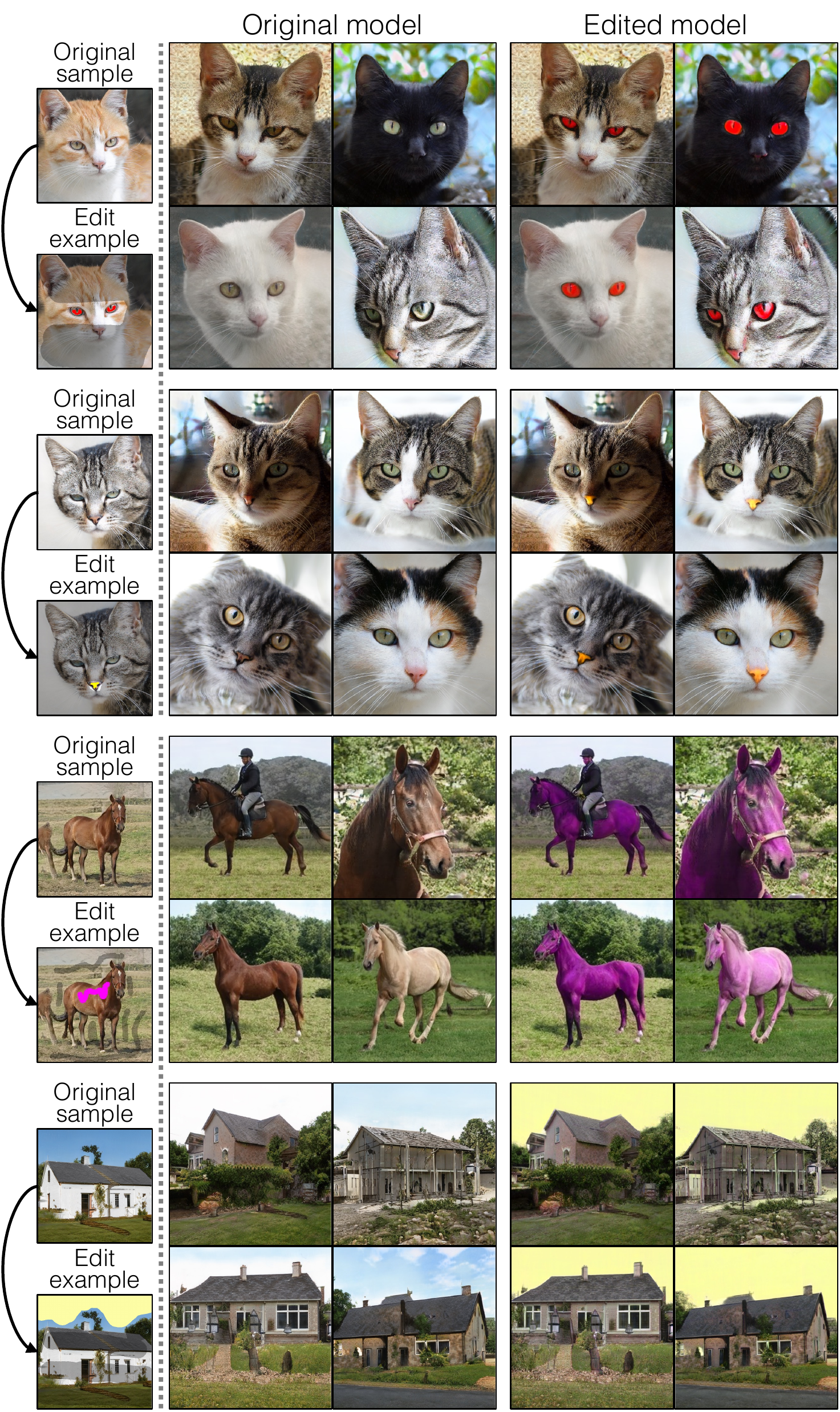}
    \caption{{\bf Color edits.} We show results of models edited with coloring operations. The first column shows the user edits. The colored strokes specify the locations to perform coloring changes, while the darker region defines the region to be preserved. The edited models produce precise coloring changes in the specified parts. In the first example (red eye), the pupil color and the flashes of light on the eye are preserved after the edits.}
    \label{fig:color_edit}
\end{figure}
\begin{figure}
    \centering
    \includegraphics[width=\linewidth]{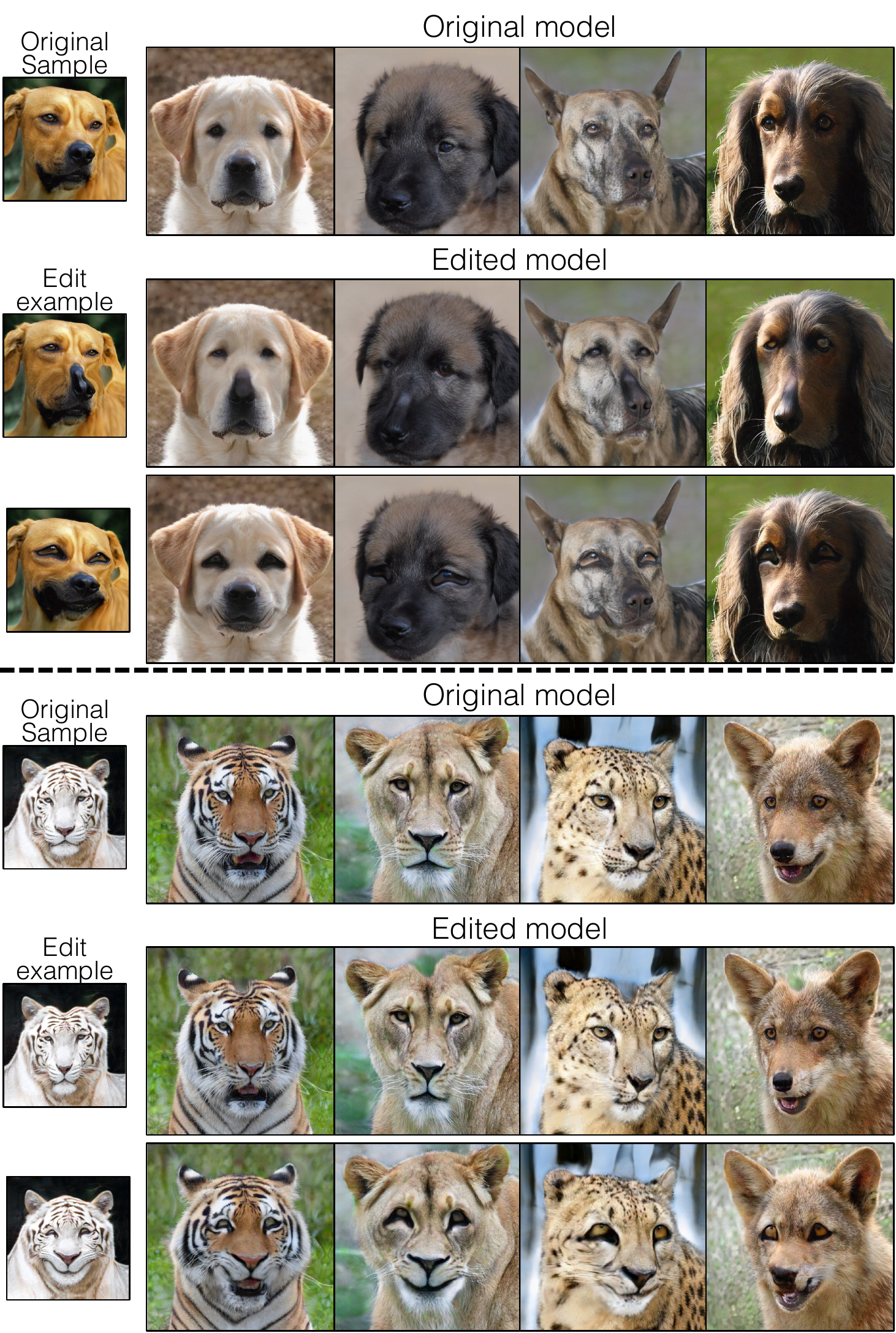}
    \caption{{\bf Qualitative results on AFHQ Dogs and Wilds.} Our method can be applied effectively on the AFHQ dog and wild classes.}
    \label{fig:afhq_extra}
\end{figure} %
\begin{figure}
    \centering
    \includegraphics[width=\linewidth]{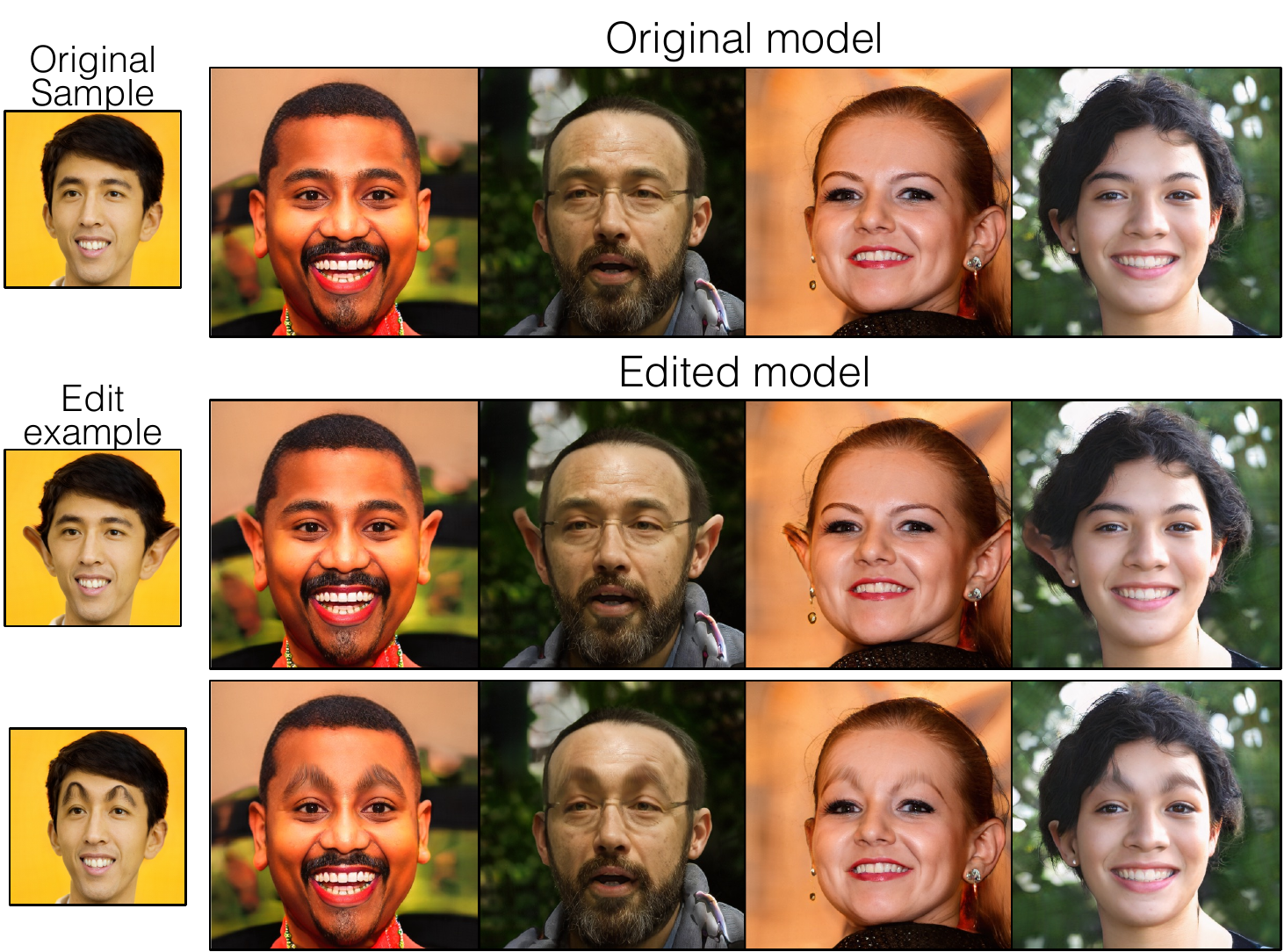}
    \caption{\camready{{\bf Qualitative results on FFHQ.} Our method can be applied effectively on the 1024px FFHQ faces.}}
    \label{fig:ffhq}
\end{figure} %
\begin{figure}
    \centering
    \includegraphics[width=\linewidth]{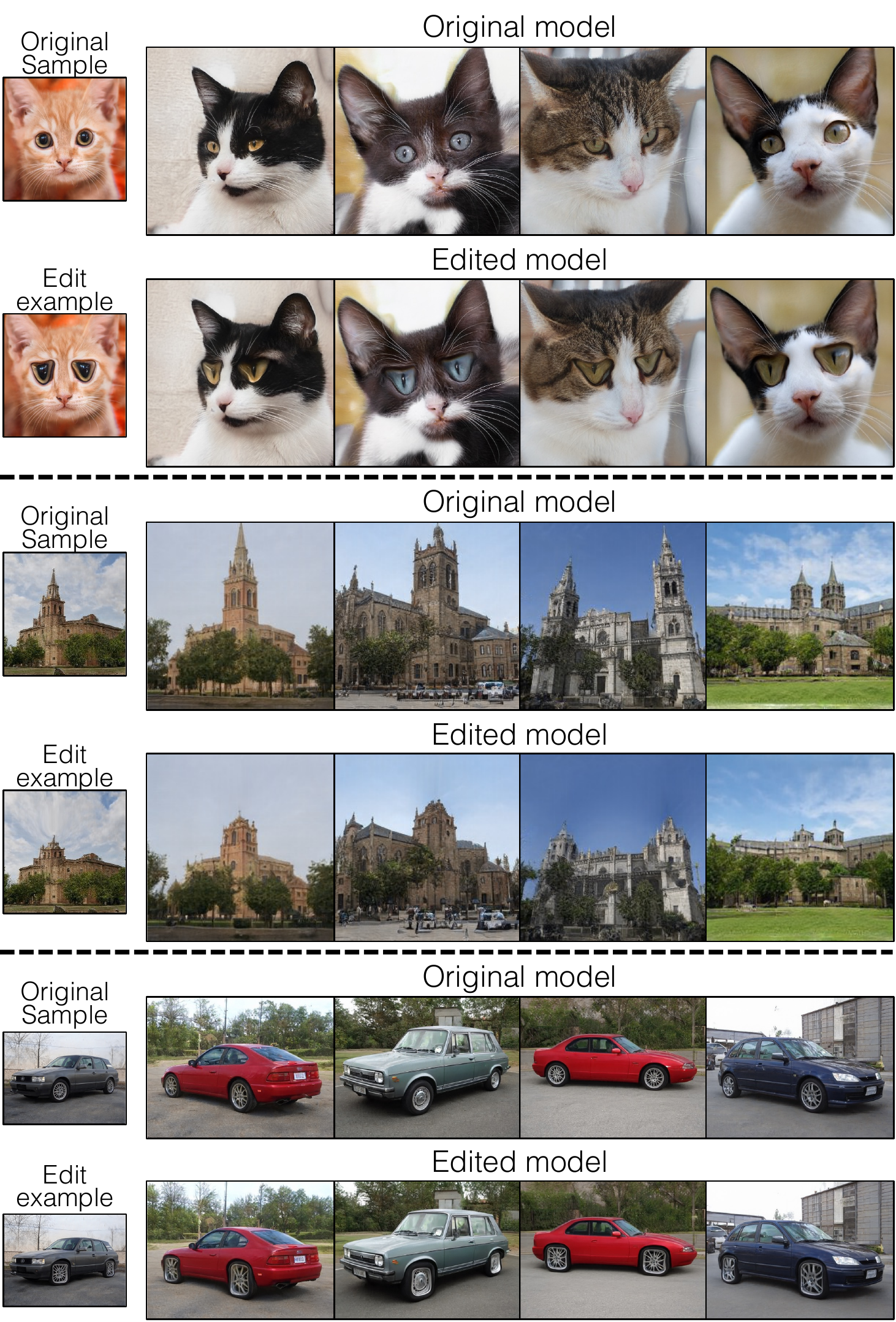}
    \caption{\camready{{\bf Qualitative results on StyleGAN2 models.} Our method can be applied effectively on StyleGAN2~\cite{karras2020analyzing}. From top to bottom, we alter the eye shapes for a AFHQ cat model, lower the tower height for a LSUN church model, and turn the tires into squares for a LSUN car model.}}
    \label{fig:sg2}
\end{figure} %
\begin{figure}
    \centering
    \includegraphics[width=\linewidth]{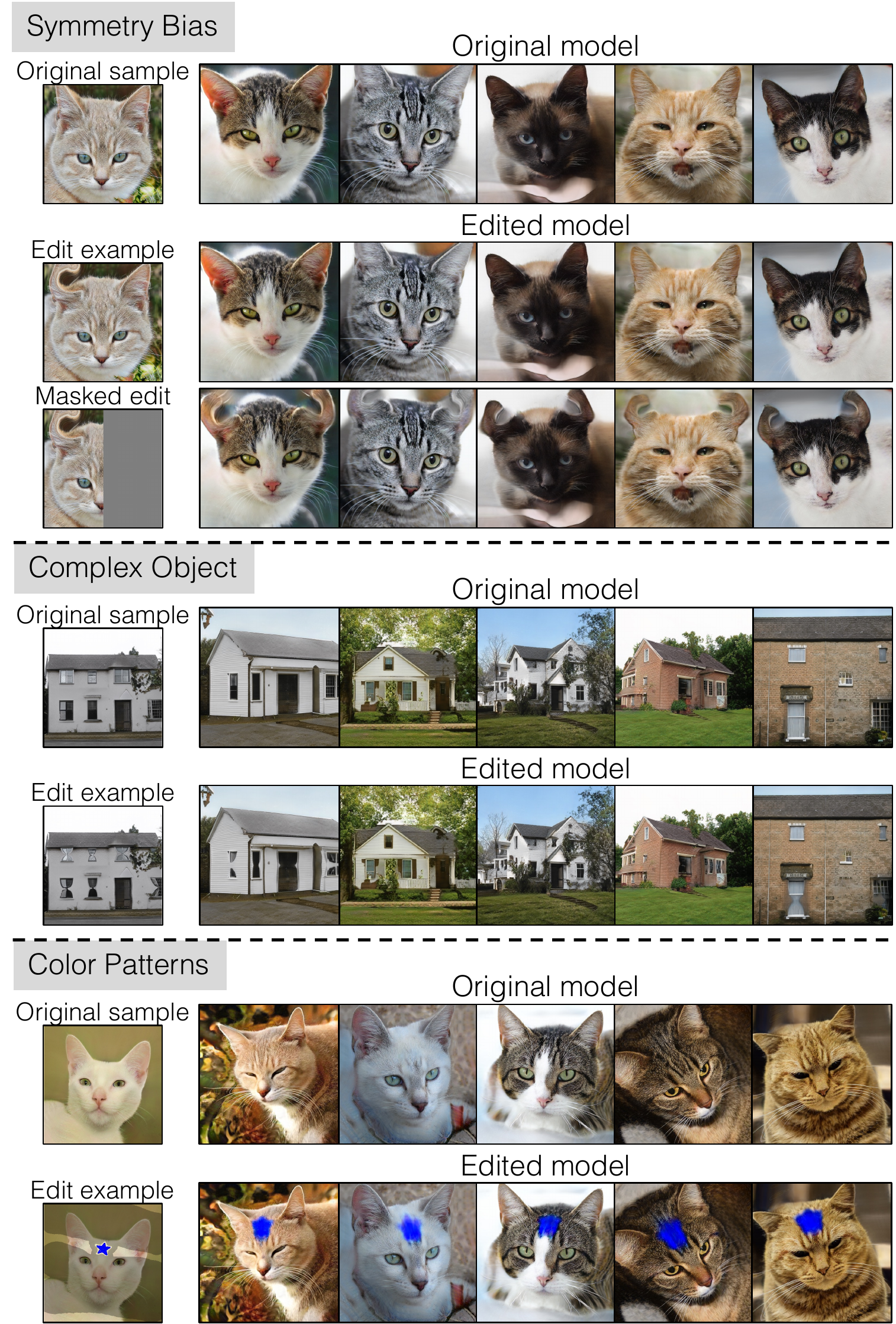}
    \caption{{\bf Failure cases.} (Top) Our method cannot apply asymmetric shape edits to a cat model, such as warping just a cat's left ear. Next, we mask out the right halves of the same images to train the model. The second model does change the shape of both ears, without seeing any right ears. (Middle) Our method cannot edit a house model to produce consistent shape changes to the windows. We observe that not every windows in the generated samples undergo the desired shape changes. \camready{(Bottom) Our method does not work well with inserting new objects, such as a “star sticker” onto cats' foreheads. Although the stars in the edited models do appear at the correct location, the results are blurry.}}
    \label{fig:failure}
\end{figure} \paragraph{Edit models with color strokes.}
As discussed in~\refsec{color}, we can also edit a model to match the user-specified color edits. As shown in~\reffig{color_edit}, the color edits are faithfully transferred to the corresponding parts in the unseen samples. Interestingly, when we change the color of cat eyes to red, the pupil color and the flashes of light on the eye remain intact. \camready{Furthermore, color edits can be applied to different models, such as horses and houses.} This effect can be attributed to the disentangled representation learned in the source models.

\paragraph{\camready{Additional results.}}
\camready{We show qualitative results on AFHQ dog and wild (\reffig{afhq_extra}), FFHQ faces (\reffig{ffhq}), and StyleGAN2~\cite{karras2020analyzing} (\reffig{sg2}). All models are trained with ten user edits.}
\section{Discussion and Limitations}
\label{sec:discussion}
We have presented a method for altering the local geometric rules of a pre-trained GAN model. 
With our method, a user can provide just a handful of control points as warping instructions to edit a pre-trained model. The edited model can faithfully synthesize endless samples that follow the desired geometric changes. Surprisingly, the geometric changes do not need to be within the original data distribution, so users have the freedom to customize their models and introduce a wide range of shapes. In addition, our models with different shape edits can be composed together to achieve the new effects, and we present an interactive interface for users to create new models by composing a preset of edited models.

One natural question is to understand the range of geometric changes where we can successfully apply our method. In~\reffig{failure}, we present several test cases to study this. In the first example, we train a model on ten warping examples where only the left ear is warped, and we find that the resulting model cannot produce the desired shape changes. In the second example, we train a model on the same set of examples, but this time the supervision only comes from the left halves of the images. This is achieved by masking out the right halves for both the model output and training images when calculating the loss. Interestingly, without seeing any examples of right ears, the resulting model applies the geometric change to the right ears as well. This suggests that the pre-trained cat model has a symmetry bias, so it is difficult to perform asymmetric shape changes to the edited models. 

Moreover, in the house example, we test if we can warp the windows into the shape of hourglasses in the house model, and we observe that only a small number of windows change their shape accordingly in the edited model, potentially due to the large style and shape variations of windows. %
\camready{Also, we find that our method fails to add new objects. We experimented with adding a “star sticker” onto cats' foreheads. The stars appear at the correct locations in the edited model, but the results are blurry.}

Additionally, our work focuses on keypoint-based warping. It would be great to support other types of warping algorithms such as mesh-based Liquify~\cite{kim2019facial} and line-based methods~\cite{beier1992feature}, as well as semi-automatic methods~\cite{aberman2018neural}. Finally, we would like to further reduce our current training time (20-40 minutes). One possible solution is to pre-compute a shared warping basis in a certain layer that could work for different types of geometric changes.  %

\paragraph{Societal impact.}
\camready{While our work makes it easier for users to create generative models and image content, potential misuse of this method unfortunately exists. For instance, a malicious user could apply facial expression changes to a face generative model to spread misinformation. These manipulations could change the narrative, for example editing a frowning political figure to be smiling, and it is important to develop methods to counter such abuses. One potential solution is to watermark the content whenever it is generated using this method. Another way is to train a classifier to detect whether an image is generated by a generative model or not~\cite{rossler2019faceforensics++,zhang2019detecting,wang2020cnn,frank2020leveraging}. We test a standard CNN-detection method~\cite{wang2020cnn} on our warped cat models along with real AFHQv2 cat images. The classifier gets 98.3\% average precision but 59.58\% accuracy. As suggested by~\cite{wang2020cnn}, the classifier trained only with ProGAN~\cite{karras2018progressive} images obtains correct rankings of real-verse-fake, but it would require calibration using a few samples. A classifier that achieves high fake-detection accuracy for new, unseen models remains an open challenge.}

\begin{acks}
\camready{We thank Richard Zhang, Nupur Kumari, Gaurav Parmar, George Cazenavette for the helpful discussions. We thank Nupur Kumari and George Cazenavette again for the significant help with proof reading and writing suggestions. We are grateful to Ruihan Gao for the edit examples in~\reffig{afhq_extra}. We truly appreciate that Flower, Sheng-Yu's sister's cat, agreed to have her portrait edited in~\reffig{real_image_edit}.  S.-Y. Wang is partly supported by a Uber Presidential Fellowship. The work is partly supported by Adobe Inc. and Naver Corporation. }
\end{acks}

\bibliographystyle{ACM-Reference-Format}
\bibliography{main}

\setcounter{section}{0}
\renewcommand\thesection{\Alph{section}}
\renewcommand{\thefootnote}{\arabic{footnote}}

\clearpage
\noindent{\huge \bf Appendix}
\begin{table}
\caption{{\bf Few-shot GANs trained with style-mixing augmentation.} We compare our rank-50 single layer update method against few-shot GANs trained with style-mixing augmentation. We find that our method still outperforms significantly.}
{\small
    \centering
    \resizebox{1.\linewidth}{!}{
\begin{tabular}{clccccccc}
\toprule
\multirow{3}{*}{Class} & \multirow{3}{*}{Name} & \multicolumn{4}{c}{Edited Region} & \multicolumn{3}{c}{Unaltered region} \\ \cmidrule(lr){3-6} \cmidrule(lr){7-9} 
 &  & PSNR & SSIM & LPIPS & Chamfer & PSNR & SSIM & LPIPS \\
 &  & ($\uparrow$) & ($\uparrow$) & ($\downarrow$) & ($\downarrow$) & ($\uparrow$) & ($\uparrow$) & ($\downarrow$) \\ \midrule \midrule
\multirow{5}{*}{Cat} & TGAN & 9.06 & 0.25 & 0.47 & 6.69 & 9.81 & 0.47 & 0.49 \\
 & TGAN+ADA & 9.58 & 0.26 & 0.41 & 5.84 & 10.81 & 0.46 & 0.42 \\
 & FreezeD & 9.29 & 0.26 & 0.45 & 6.73 & 9.99 & 0.48 & 0.48 \\
 & \multicolumn{1}{l}{\cite{ojha2021few-shot-gan}} & 8.73 & 0.24 & 0.47 & 7.32 & 10.27 & 0.44 & 0.43 \\ \cmidrule(l{5pt}r{5pt}){2-9} 
 & Ours & \textbf{11.83} & \textbf{0.36} & \textbf{0.30} & \textbf{5.07} & \textbf{21.09} & \textbf{0.77} & \textbf{0.08} \\ \midrule
\multirow{5}{*}{Horse} & TGAN & 9.14 & 0.20 & 0.41 & 10.19 & 9.36 & 0.19 & 0.40 \\
 & TGAN+ADA & 9.30 & 0.18 & 0.41 & 10.49 & 9.54 & 0.19 & 0.39 \\
 & FreezeD & 9.78 & 0.21 & 0.40 & 7.48 & 9.57 & 0.19 & 0.40 \\
 & \multicolumn{1}{l}{\cite{ojha2021few-shot-gan}} & 7.95 & 0.16 & 0.48 & 5.86 & 9.44 & 0.20 & 0.39 \\ \cmidrule(l{5pt}r{5pt}){2-9} 
 & Ours & \textbf{11.98} & \textbf{0.28} & \textbf{0.30} & \textbf{3.87} & \textbf{20.41} & \textbf{0.79} & \textbf{0.05} \\ \midrule
\multirow{5}{*}{House} & TGAN & 6.60 & 0.28 & 0.59 & 10.05 & 7.73 & 0.32 & 0.54 \\
 & TGAN+ADA & 6.86 & 0.29 & 0.58 & 9.68 & 8.37 & 0.33 & 0.51 \\
 & FreezeD & 6.78 & 0.26 & 0.57 & 8.79 & 7.41 & 0.29 & 0.52 \\
 & \multicolumn{1}{l}{\cite{ojha2021few-shot-gan}} & 5.44 & 0.30 & 0.70 & 11.71 & 8.45 & 0.34 & 0.53 \\ \cmidrule(l{5pt}r{5pt}){2-9} 
 & Ours & \textbf{10.04} & \textbf{0.37} & \textbf{0.37} & \textbf{5.19} & \textbf{16.45} & \textbf{0.60} & \textbf{0.11} \\ \bottomrule
\end{tabular}
}}
\label{tbl:fewshot_stymix}
\end{table} %
\begin{figure}
    \centering
    \includegraphics[width=\linewidth]{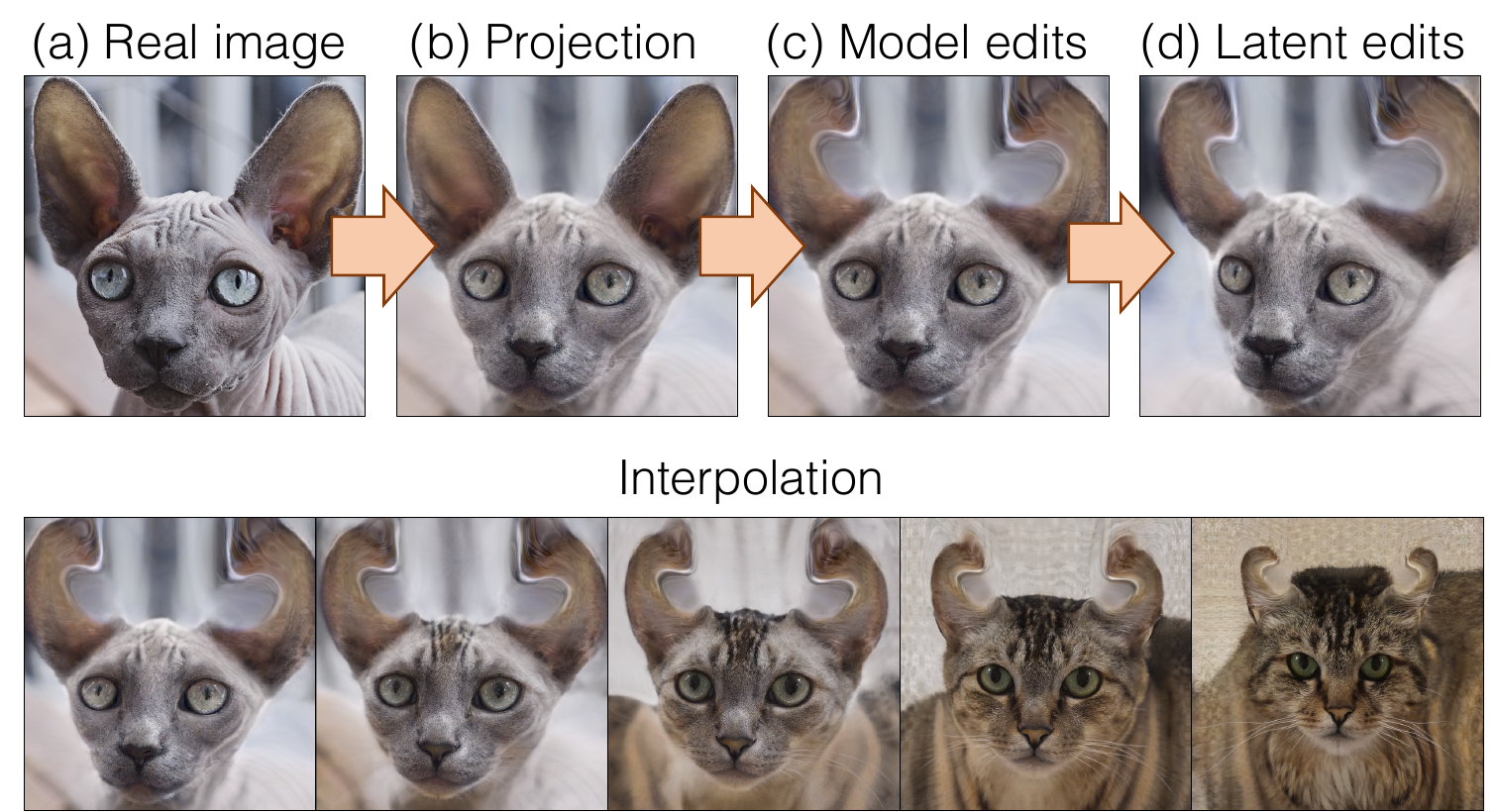}
    \caption{{\bf Sphynx cat example.} We show that our method can be applied to a Sphynx cat, which has a unique color and shape. (Top) the cat image can be edited the same way as Figure 13 in the main text. (Bottom) The projected cat sample interpolates smoothly to a random sample using our model.}
    \label{fig:sphynx}
\end{figure}

\begin{table}
    \caption{{\bf StyleGAN-NADA text prompts.} The prompts are used to evaluate one of the baselines (\textbf{NADA(Text)}). We note that StyleGAN-NADA~\cite{gal2021stylegan} uses a source-to-target text prompt pair to update a generative model.}
    \centering
    {%
    \centering
    \begin{tabular}{ll}
        \toprule
        Source & Target  \\ \midrule \midrule
        shape of a cat face & shape of a rabbit face \\
        shape of cat ears & shape of devil ears \\
        shape of cat eyes & shape of alien eyes \\
        normal-shaped cat nose & cross-shaped cat nose \\
        shape of a horse back & shape of a camel back \\
        a horse standing & a horse with a lifted front leg \\
        shape of a house roof & shape of a curly house roof \\
        shape of a normal ground & shape of a risen ground \\ \bottomrule
    \end{tabular}}
    \label{tbl:nada_prompt}
\end{table} 

\section{Additional Analysis}
\myparagraph{Few-shot GANs trained with style-mixing augmentation.}
We test several few-shot GAN methods trained on the edited examples with style-mixing augmentation applied, and results are shown in~\reftbl{fewshot_stymix}. Comparing to Table 1 in the main text, we find that style-mixing augmentation improves most baselines, but the performance remains significantly inferior to our full method. This is because that few-shot GANs do not explicitly enforce correspondence between original samples and edited samples, even if style-mixing augmentation improves the diversity of the training set.

\myparagraph{Runtime comparison to StyleGAN-NADA.} 
According to Gal~\etal~\shortcite{gal2021stylegan}, StyleGAN-NADA’s training time depends on the edit types. While texture-based changes take 1-3 minutes on a V100, complex shape edits can take up to 6 hours.

Further, we observe that training \emph{NADA (image)} and \emph{NADA (text)} with longer iterations does not improve the performance in all of our tasks. We train StyleGAN-NADA with 10x more iterations than the default iteration counts used in our paper, ~30 minutes on an A5000. For cat models, when trained longer, \emph{NADA (image)} gets 6.60 and 9.11 in PSNR for edited and unaltered regions, respectively. This is worse than \emph{NADA (image)}’s performance reported in the paper (8.13, 14.20), and our method’s performance (11.83, 21.09).

\begin{figure}
    \centering
    \includegraphics[width=\linewidth]{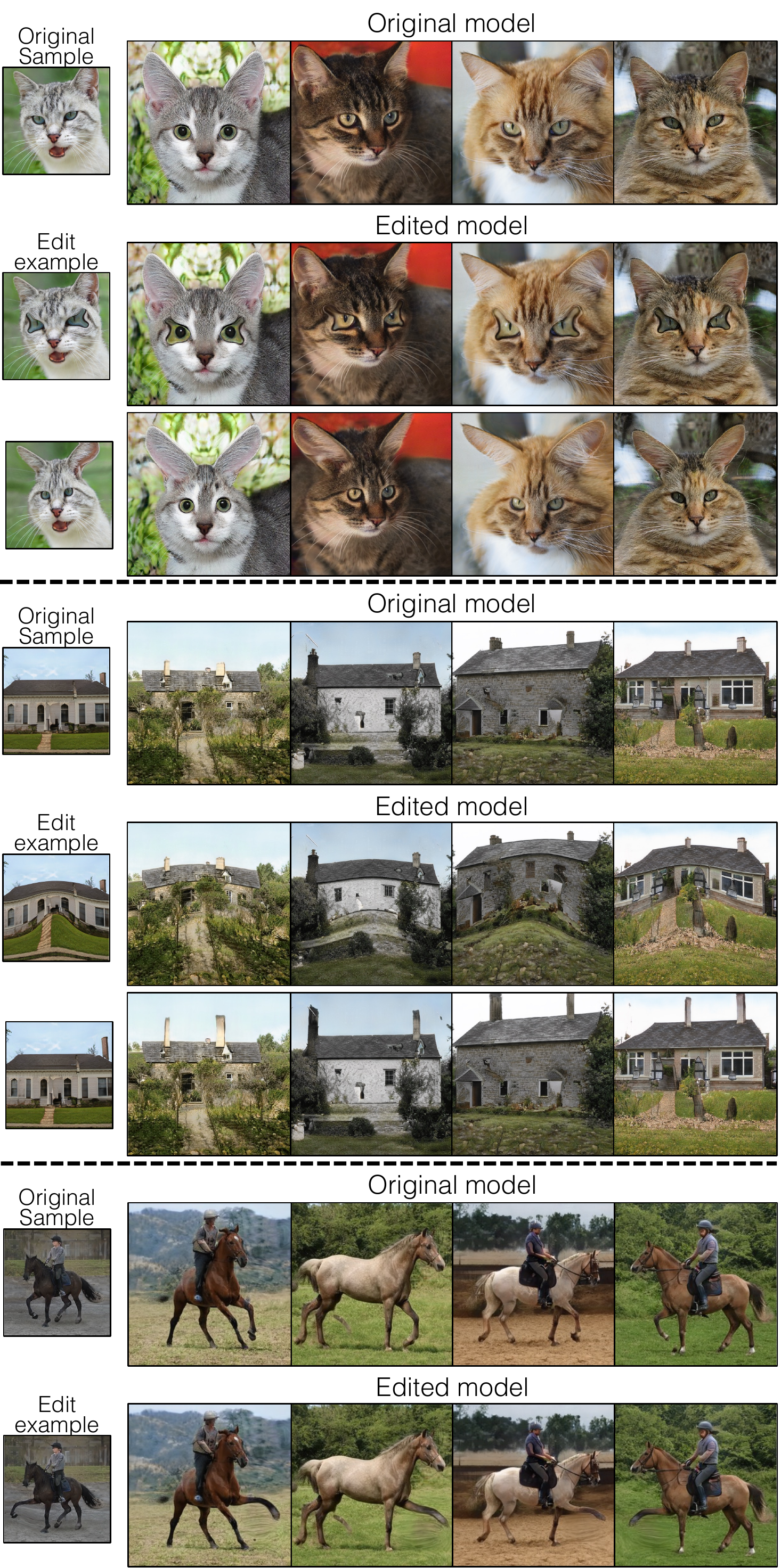}
    \caption{{\bf Extra qualitative results.} We show more model warping results on StyleGAN3 cat, house, and horse models.}
    \label{fig:extra_qual}
\end{figure}

\myparagraph{Additional qualitative results.}
Our geometric edits generalize well to new latent codes with color, texture, shape, and pose variations. As shown in~\reffig{sphynx}, we tested our method on latent codes that represent new shapes -- a sample projected from a real Sphynx cat image. Our method succeeded in changing the eyes and ears of the Sphynx cat and producing smooth interpolations between the Sphynx cat latent code and other latent codes. We also include additional model warping results in~\reffig{extra_qual}.

\section{Implementation Details}
\myparagraph{Details of the Chamfer Distance metric.}
For each pair of hold out ground truth edit example and edited model's output, we predict the contour of both images using Dexined pre-trained on BIPEDv2~\cite{xsoria2020dexined}. The model can predict detailed contours, which helps us better evaluate fine-grained parts like eyes or noses. We then post-process the contour following Isola et al.~\shortcite{isola2017image} to obtain a single-pixel-wide contour map. Finally, we calculate the symmetric Chamfer distance on the edited region. %

\myparagraph{Text prompts used for StyleGAN-NADA.}
\reftbl{nada_prompt} shows the text prompts used to evaluate the baseline \emph{NADA (text)}. StyleGAN-NADA~\cite{gal2021stylegan} finetunes a GAN using a source-to-target text prompt. The GAN is fine-tuned using the difference in the CLIP embedding between the text pair.

\myparagraph{Model Rewriting implementation.}
To apply Model Rewriting~\cite{bau2020rewriting} to our warping tasks, for each training sample, we manually mask the region that will be warped. The masks are used to compute and fix the context key for the proposed rank-1 update method. We optimize for the reconstruction loss between the original and edited samples to obtain the edited model. We also test a variation of the baseline using rank-50 updates.

\myparagraph{CATs implementation.}
We apply CATs~\cite{cho2021cats} to predict the correspondence between a given test sample and the 10 training samples. For each training sample, we transform the warping field using the estimated correspondence. In total, we get 10 transformed warping fields for each test sample, and we use the averaged warping field to warp the test sample.

\myparagraph{Hyperparameter details}
We show the hyperparameters of warped StyleGAN3 models, color-edited StyleGAN3 models, and warped StyleGAN2 models in \reftbl{sg3_warp_hyper}, \reftbl{sg3_color_hyper}, and \reftbl{sg2_warp_hyper}, respectively.

\clearpage

\begin{table}[h]
    \caption{{\bf Warped StyleGAN3 hyperparameters.} We show the layer used for training (Layer), learning rate (Lr), and whether it is evaluated quantitatively in the paper (Evaluated). Default hyperparameters are used if not mentioned.}
    {%
    \centering
    \begin{tabular}{clccc}
        \toprule
        \multirow{2}{*}{Class} & \multirow{2}{*}{Warp type} & \multicolumn{3}{c}{Details}   \\ \cmidrule(lr){3-5}
        & & Layer & Lr & Evaluated
        \\ \midrule \midrule
        \multirow{4}{*}{Cat} & Devilish ears & 8 & 0.05 & $\checkmark$  \\
        & Cross-shaped noses & 8 & 0.05 & $\checkmark$  \\
        & Alien eyes & 8 & 0.05 & $\checkmark$  \\
        & Rabbit faces & 1 & 0.05 & $\checkmark$  \\ \midrule
        \multirow{2}{*}{Horse} & Camel backs & 8 & 0.05 & $\checkmark$  \\
        & Lifted front leg & 8 & 0.05 & $\checkmark$  \\ \midrule
        \multirow{3}{*}{House} & Curly roofs & 8 & 0.05 & $\checkmark$  \\
        & Heaved ground & 1 & 0.05 & $\checkmark$  \\
        & Taller chimneys & 8 & 0.05 &   \\ \midrule
        \multirow{2}{*}{FFHQ} & Elf ears & 5 & 0.01 & \\
        & Arched eyebrows & 6 & 0.01 & \\
\bottomrule
    \end{tabular}}
    \label{tbl:sg3_warp_hyper}
\end{table}

\begin{table}[h]
    \caption{{\bf Color-edited StyleGAN3 hyperparameters.} We show the layer used for training (Layer) and the weight of the color loss ($\lcolor$). Default hyperparameters are used if not mentioned.}
     {%
    \centering
    \begin{tabular}{clcc}
        \toprule
        \multirow{2}{*}{Class} & \multirow{2}{*}{Color-edit type} & \multicolumn{2}{c}{Details}   \\ \cmidrule(lr){3-4}
        & & Layer & $\lcolor$
        \\ \midrule \midrule
        \multirow{3}{*}{Cat} & Red eyes & 13 & 8  \\
        & Golden noses & 13 & 8  \\
        & Blue star & 10 & 8  \\ \midrule
        \multirow{1}{*}{Horse} & Purple fur & 14 & 1  \\ \midrule
        \multirow{1}{*}{House} & Yellow sky & 13 & 8  \\
    \bottomrule
    \end{tabular}}
    \label{tbl:sg3_color_hyper}
\end{table}

\begin{table}[h]
    \caption{{\bf Warped StyleGAN2 hyperparameters.} We show the layer used for training (Layer) and the learning rate (Lr). Default hyperparameters are used if not mentioned.}
     {%
    \centering
    \begin{tabular}{clcc}
        \toprule
        \multirow{2}{*}{Class} & \multirow{2}{*}{Warp type} & \multicolumn{2}{c}{Details}   \\ \cmidrule(lr){3-4}
        & & Layer & Lr
        \\ \midrule \midrule
        AFHQ Cat & Big triangular eyes & 8 & 0.05  \\
        LSUN Church & Shorter towers & 4 & 0.05  \\
        LSUN Car & Square-shaped tires & 8  & 0.05  \\
    \bottomrule
    \end{tabular}}
    \label{tbl:sg2_warp_hyper}
\end{table}

\end{document}